%% file: main.tex
\documentclass[12pt]{article}

\usepackage{fullpage}

\usepackage{amsmath,amsfonts}

\usepackage[pdftex]{graphicx}

\usepackage{adjustbox}
\usepackage{tipa}
\usepackage{catchfilebetweentags}
\usepackage{hyperref}

\usepackage{microtype}

\usepackage{tikz}
\usepackage{setspace}
\usepackage{authblk}

\usepackage[square,numbers]{natbib} 

\usepackage{times}

\begin{document}


\title{Multiple evolutionary pressures shape 
identical consonant avoidance in the world's languages}

\author[a,b,c]{Chundra A. Cathcart}

\affil[a]{Department of Comparative Language Science, University of Zurich}
\affil[b]{Center for the Interdisciplinary Study of Language Evolution, University of Zurich}
\affil[c]{DFG Center ``Words, Bones, Genes, Tools'', University of T\"{u}bingen}



\maketitle

\input{abstract}

\onehalfspacing



\section{Introduction}

\input{introduction}

\section{Results}

\input{results}

\input{discussion}


\section{Materials and Methods}

\input{methods}





\section*{Acknowledgements}

\input{acknowledgements}

\bibliographystyle{unsrt}
\bibliography{bibliography}

\end{document}

%% file: abstract.tex
\begin{abstract}

\noindent Languages disfavor word forms containing sequences of similar or identical consonants, due to the biomechanical and cognitive difficulties posed by patterns of this sort. 
However, the specific evolutionary processes responsible for this phenomenon are not fully understood. 
Words containing sequences of identical consonants may be more likely to arise than those without; processes of word form mutation may be more likely to remove than create sequences of identical consonants in word forms; 
finally, words containing identical consonants may die out more frequently than those without. 
Phylogenetic analyses of the evolution of 
homologous word forms 
indicate that words with identical consonants arise less frequently than those without, and processes which mutate word forms are more likely to remove sequences of identical consonants than introduce them. 
However, words with identical consonants do not die out more frequently than those without. 
Further analyses reveal that forms with identical consonants 
are replaced in basic meaning functions more frequently than words without. 
Taken together, results suggest that the under-representation of sequences of identical consonants is overwhelmingly a byproduct of constraints on word form coinage, though processes related to word usage also serve to ensure that such patterns are infrequent in more salient vocabulary items. 
These findings clarify previously unknown aspects of processes of lexical evolution and competition that take place during language change, optimizing communicative systems.


\end{abstract}

%% file: introduction.tex
The world's spoken languages vary considerably in terms of the combinations of sounds they allow within words, as well as the frequencies of different static sound patterns they display. 
For instance, a hypothetical word {\it bnick} is not well formed in English, but similar forms are valid in other languages, e.g., Moroccan Arabic {\it bniqa} `closet' \cite{ChomskyHalle1968,gorman2011program}. Preferences for specific patterns of this sort are highly stable within groups of closely related languages \cite{macklin2021phylogenetic}. 
At the same time, a number of quasi-universal patterns have been identified in large numbers of genetically diverse languages with respect to the sound patterns they display \cite{blasi2016sound,johansson2020typology,everett2018global}. 
One such phenomenon is the statistical under-representation of proximate similar or identical consonants within lexical items, documented in a diverse sample of languages: 
all else being equal, a sequence of identical consonants separated by a vowel 
is far less likely to be found in the vocabularies of the world's languages than would be expected according to chance. 
In some language-specific cases, restrictions on such sequences are categorical: Arabic contains no words in which the first two consonants are identical, such as the hypothetical form {\it sasam} \cite{greenberg1950patterning,mccarthy1986ocp}. 

Constraints 
on similar or identical adjacent elements are documented at a number of linguistic levels \cite{nevins2012haplological}. 
Repetition of formally identical elements tends to be dispreferred within words (e.g., {\it sillily}, {\it friendlily} etc., are deemed unacceptable by many English speakers). 
Additionally, some languages do not allow identical case markers to appear on adjacent words \cite{allen1984certain}, exhibiting identical element avoidance at the sentence level. 
The extent to which similarity avoidance obtains in non-human communication systems has not been fully investigated. 
Call 
sequences in Putty-nosed monkeys reveal a high degree of tolerance for adjacent identical elements \cite{arnold2012call,schlenker2016pyow}. 
However, while Black-and-white Colobus monkeys can produce adjacent roars, adjacent snorts cannot occur without intervening pauses \cite{schel2009alarm}, suggesting the existence of constraints on certain sequences of identical call types. 

The avoidance of similar and identical adjacent consonants is reported in a variety of languages from different 
language families 
\cite{buckley1997tigrinya,berkley2000gradient,Frischetal2004,pozdniakov2007similar,coetzee2008weighted,WilsonObdeyn2009,graff2009locality,racz2016gradient,grotberg2022quantifying,stanton2022defense}. 
Experimental findings indicate that forms containing identical consonants are difficult to process and produce.  
Participants in lexical decision tasks are slower to recognize words and faster to reject non-words containing identical consonants \cite{vandeWeijer2005}. 
Listeners are less likely to perceive ambiguous synthesized stimuli as containing identical consonants \cite{coetzee2008grammaticality}. 
Utterance onset times occur later for words containing similar sounds than for words containing dissimilar sounds in production experiments \cite{COHENGOLDBERG2012184}. 
Additionally, while repeated syllables are easier for children to produce and learn \cite{ota2016reduplicated}, adults exhibit a faster speech rate for sequences of different syllables \cite{lancheros2020neural}, striking given that 
identical consonants are common in nursery words 
(e.g., {\it mama}, {\it cookie}, etc.). 

\begin{figure}
\centering

\small{

\frame{

\input{cognate-figure}

}

\frame{

\input{cognate-concept-figure}

}

}
\caption{Schemata of 
continuous-time Markov models of evolution for a cognate class trait representing the Proto-Malayo-Polynesian etymon *dapdap (above) 
and a cognate-concept trait representing whether languages use the Proto-Indo-European root {*peh$_3$-} in the meaning `drink' (below). 
Both trait types undergo transitions between states representing absence, presence without identical consonants, and presence with identical consonants. 
Tree branch colors represent hypothetical but unobserved character histories (i.e., evolutionary trajectories) involving transitions between states. 
Transition rates (representing frequencies of transitions between states) can be inferred on the basis of (1) data attested in languages and (2) language phylogenies. 
Parameters governing the evolution of the traits given here can be subdivided into birth rates $\lambda^{-}_0$, $\lambda^{+}_0$ (transitions from {\sc absent} to $\pm$IC), rates involving mutations introducing or removing sequences of identical consonants $\rho^{-+}_0$, $\rho^{+-}_0$ (transitions between $\pm$IC), and loss rates $\mu^{-}_0$, $\mu^{+}_0$ involving the death of cognate classes or concept-cognate traits (transitions from $\pm$IC to {\sc absent}). The dashed lines in the schema in the top panel represent the understanding that cognate classes are born only once.}
\label{fig:ctm}
\end{figure}
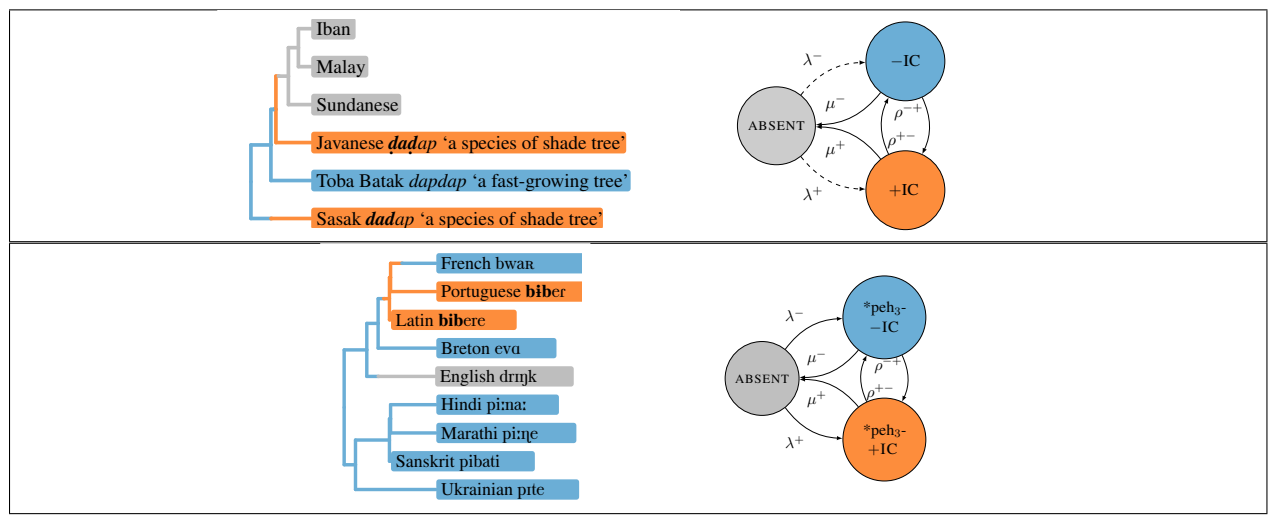

Despite the well-documented nature of this phenomenon 
and ample experimental evidence that avoidance of this type is beneficial for both word production and comprehension, very little is known about the specific diachronic mechanisms involved in the emergence and maintenance of this pattern. 
There are several orthogonal processes of language change that may exert pressure on linguistic systems to disfavor word forms containing identical consonants, but the role of these different mechanisms remains unexplored. 
One possibility is that words containing identical consonants arise in languages with low frequency, 
seldom coined or borrowed from other speech varieties. 
Speakers of languages are often incapable of creating or borrowing words that do not adhere to the static sound patterns to which they are habituated, and will often adapt word forms to these patterns 
\cite{lahiri2019pertinacity}, e.g., Japanese {\it dorama} from English {\it drama}. 
Given the large body of psycholinguistic evidence that words containing identical consonants are more difficult to produce and process than those without, it may be the case that they are less likely to enter a language's vocabulary, and that when a word form arises on a phylogenetic lineage, it is unlikely to contain a sequence of identical consonants.

A second possibility is that mutational processes such as sound changes and analogical changes operating upon the phonological form of a word frequently remove sequences of identical consonants when present, and rarely introduce them when they are absent. 
In some cases, these developments give rise to sequences of identical consonants within words;
for instance, a regular sound change involving consonant cluster simplification gave rise to 
Sundanese {\it dedek} `rice bran', which descends from earlier Proto-Malayo-Polynesian {*dekdek}. 
The first consonant of the form directly ancestral to Latin {\it bibet} `he/she drinks' was most likely {\it p-} (cf.\ Sanskrit {\it pibati}, with the same meaning; both forms descend from 
a reduplicated Proto-Indo-European verb form {*pi-ph$_3$-e-ti}), but changed to {\it b-}, most likely on analogy with other reduplicated verbs for which the consonant of the reduplicant matches the consonant of the base. 
Mutations are also capable of removing sequences of identical consonants within word forms. 
While developments specifically dedicated to removing sequences of identical consonants 
are infrequent in surveys of sound changes \cite{Kuemmel2007,pozdniakov2007similar}, 
more general sound changes may 
be responsible for the rare occurrence of such sequences (e.g., Latin {\it bibere} `drink' developed to Old French {\it beivre} due to a general weakening of word-medial {\it -b-} to {\it -v-} that affected other forms, not just those containing two instances of {\it b}). 
Additionally, sporadic or irregular sound changes can operate under some circumstances in order to facilitate efficient communication \cite{blevins2009inhibited}. 
Ultimately, it is possible that a variety of processes (regular sound change, sporadic sound change, analogical change) play a role in ensuring that sequences of identical consonants arise infrequently, or are removed when they arise.

A third view found in the literature but untested empirically on a large scale hypothesizes that words 
containing sequences of identical consonants 
are rare 
due to dynamics of lexical replacement. 
While a number of processes may lead to the presence of identical consonants in a word, a lexical item may be phased out of use relatively rapidly once such a pattern arises in it, losing ground to competitor forms that do not contain the same disfavored sound pattern  \cite{Frischetal2004,martin2007evolving,pozdniakov2007similar}. 
This view of lexical change invites clear analogies with notions of selectional pressure found in biology, a force invoked in previous work on lexical evolution \cite{Pagel2007}: while a number of processes may give rise to variation in forms corresponding to a given meaning, language users will select against forms that are deleterious from the perspective of language production and processing. 

Finally, more than one of the three pressures identified above may be involved in persistence of identical consonant avoidance. 
Phylogenetic comparative methods (Figure \ref{fig:ctm}) were used to model the evolutionary dynamics of cognate class (i.e., homologous, etymologically related words that share a common origin but may differ in meaning) as well as cognate-concept traits (i.e., features which register whether a language uses a cognate class in a particular meaning function, alternatively referred to as root-meaning traits) in a diverse sample of the world's language families. 
Analyzing both of these data types sheds light on complementary dynamics of lexical evolution. On one hand, cognate classes provide a picture of the full evolutionary trajectory of homologous formal elements across related languages, but do not provide explicit information regarding lexical competition and replacement: a word form may die out conceivably because it loses out to a competitor, but as it is challenging to pinpoint the specific semantic function in which the word served before dying out, we have no information about the form that came to replace it. 
Cognate-concept traits
provide an explicit way to model lexical competition and replacement in that we can explicitly track the forms that replace each other in particular meaning functions; at the same time, models based on these features do not allow us to make inferences regarding the trajectory of a form before it comes to serve in a meaning function or after it is replaced: a form may be replaced in a specific meaning function but may go on to express another more salient concept rather than die out. 
The two sets of analyses conducted are designed to disentangle the role of the three mechanisms outlined above. 

%% file: cognate-figure.tex
\begin{adjustbox}{max totalsize={.6\linewidth}{.6\linewidth},center}
\begin{minipage}{.6\linewidth}
\begin{adjustbox}{min totalsize={\linewidth}{.6\linewidth},center}


\input{cognate-tree}

\end{adjustbox}
\end{minipage}

\hspace{.05\linewidth}

\begin{minipage}{.5\linewidth}




\begin{adjustbox}{min totalsize={.65\linewidth}{.65\linewidth},center}
\input{cognate-CTM}
\end{adjustbox}

\end{minipage}
\end{adjustbox}

%% file: cognate-tree.tex
\begin{tikzpicture}[x=1pt,y=1pt]
\definecolor{fillColor}{RGB}{255,255,255}
\path[use as bounding box,fill=fillColor,fill opacity=0.00] (0,0) rectangle (289.08,144.54);
\begin{scope}
\path[clip] (  0.00,  0.00) rectangle (289.08,144.54);
\definecolor{drawColor}{RGB}{255,255,255}
\definecolor{fillColor}{RGB}{255,255,255}

\path[draw=drawColor,line width= 0.6pt,line join=round,line cap=round,fill=fillColor] (  0.00,  0.00) rectangle (289.08,144.54);
\end{scope}
\begin{scope}
\path[clip] (  8.25,  8.25) rectangle (283.58,139.04);
\definecolor{drawColor}{RGB}{255,255,255}
\definecolor{fillColor}{RGB}{255,255,255}

\path[draw=drawColor,line width= 0.6pt,line join=round,line cap=round,fill=fillColor] (  8.25,  8.25) rectangle (283.58,139.04);
\definecolor{drawColor}{RGB}{190,190,190}

\path[draw=drawColor,line width= 2.3pt,line join=round,line cap=round] ( 50.49,109.31) -- ( 58.31,109.31);

\path[draw=drawColor,line width= 2.3pt,line join=round,line cap=round] ( 50.49,133.10) -- ( 58.31,133.10);

\path[draw=drawColor,line width= 2.3pt,line join=round,line cap=round] ( 44.23, 85.54) -- ( 58.31, 85.54);
\definecolor{drawColor}{RGB}{253,141,60}

\path[draw=drawColor,line width= 2.3pt,line join=round,line cap=round] ( 36.41, 61.76) -- ( 58.31, 61.76);
\definecolor{drawColor}{RGB}{107,174,214}

\path[draw=drawColor,line width= 2.3pt,line join=round,line cap=round] ( 33.28, 37.98) -- ( 58.31, 37.98);
\definecolor{drawColor}{RGB}{253,141,60}

\path[draw=drawColor,line width= 2.3pt,line join=round,line cap=round] ( 33.28, 14.20) -- ( 58.31, 14.20);

\path[] ( 20.77, 37.23) -- ( 20.77, 37.23);
\definecolor{drawColor}{RGB}{107,174,214}

\path[draw=drawColor,line width= 2.3pt,line join=round,line cap=round] ( 20.77, 60.27) -- ( 33.28, 60.27);

\path[draw=drawColor,line width= 2.3pt,line join=round,line cap=round] ( 33.28, 82.56) -- ( 34.84, 82.56);

\path[draw=drawColor,line width= 2.3pt,line join=round,line cap=round] ( 34.84, 82.56) -- ( 36.41, 82.56);
\definecolor{drawColor}{RGB}{253,141,60}

\path[draw=drawColor,line width= 2.3pt,line join=round,line cap=round] ( 36.41,103.37) -- ( 37.97,103.37);
\definecolor{drawColor}{RGB}{190,190,190}

\path[draw=drawColor,line width= 2.3pt,line join=round,line cap=round] ( 37.97,103.37) -- ( 44.23,103.37);

\path[draw=drawColor,line width= 2.3pt,line join=round,line cap=round] ( 44.23,121.20) -- ( 50.49,121.20);
\definecolor{drawColor}{RGB}{107,174,214}

\path[draw=drawColor,line width= 2.3pt,line join=round,line cap=round] ( 20.77, 14.20) -- ( 33.28, 14.20);
\definecolor{drawColor}{RGB}{190,190,190}

\path[draw=drawColor,line width= 2.3pt,line join=round,line cap=round] ( 50.49,121.20) -- ( 50.49,109.31);

\path[draw=drawColor,line width= 2.3pt,line join=round,line cap=round] ( 50.49,121.20) -- ( 50.49,133.10);

\path[draw=drawColor,line width= 2.3pt,line join=round,line cap=round] ( 44.23,103.37) -- ( 44.23, 85.54);
\definecolor{drawColor}{RGB}{253,141,60}

\path[draw=drawColor,line width= 2.3pt,line join=round,line cap=round] ( 36.41, 82.56) -- ( 36.41, 61.76);
\definecolor{drawColor}{RGB}{107,174,214}

\path[draw=drawColor,line width= 2.3pt,line join=round,line cap=round] ( 33.28, 60.27) -- ( 33.28, 37.98);
\definecolor{drawColor}{RGB}{253,141,60}

\path[draw=drawColor,line width= 2.3pt,line join=round,line cap=round] ( 33.28, 14.20) -- ( 33.28, 14.20);

\path[] ( 20.77, 37.23) -- ( 20.77, 37.23);
\definecolor{drawColor}{RGB}{107,174,214}

\path[draw=drawColor,line width= 2.3pt,line join=round,line cap=round] ( 20.77, 37.23) -- ( 20.77, 60.27);

\path[draw=drawColor,line width= 2.3pt,line join=round,line cap=round] ( 33.28, 60.27) -- ( 33.28, 82.56);

\path[draw=drawColor,line width= 2.3pt,line join=round,line cap=round] ( 34.84, 82.56) -- ( 34.84, 82.56);
\definecolor{drawColor}{RGB}{253,141,60}

\path[draw=drawColor,line width= 2.3pt,line join=round,line cap=round] ( 36.41, 82.56) -- ( 36.41,103.37);
\definecolor{drawColor}{RGB}{190,190,190}

\path[draw=drawColor,line width= 2.3pt,line join=round,line cap=round] ( 37.97,103.37) -- ( 37.97,103.37);

\path[draw=drawColor,line width= 2.3pt,line join=round,line cap=round] ( 44.23,103.37) -- ( 44.23,121.20);
\definecolor{drawColor}{RGB}{107,174,214}

\path[draw=drawColor,line width= 2.3pt,line join=round,line cap=round] ( 20.77, 37.23) -- ( 20.77, 14.20);
\definecolor{fillColor}{RGB}{190,190,190}

\path[fill=fillColor] ( 60.30,102.50) --
	( 92.45,102.50) --
	( 92.38,102.50) --
	( 92.67,102.52) --
	( 92.96,102.57) --
	( 93.23,102.68) --
	( 93.48,102.82) --
	( 93.70,103.01) --
	( 93.90,103.22) --
	( 94.05,103.47) --
	( 94.17,103.74) --
	( 94.24,104.02) --
	( 94.26,104.31) --
	( 94.26,104.31) --
	( 94.26,114.32) --
	( 94.26,114.32) --
	( 94.24,114.61) --
	( 94.17,114.89) --
	( 94.05,115.16) --
	( 93.90,115.41) --
	( 93.70,115.62) --
	( 93.48,115.81) --
	( 93.23,115.95) --
	( 92.96,116.06) --
	( 92.67,116.11) --
	( 92.45,116.13) --
	( 60.30,116.13) --
	( 60.52,116.11) --
	( 60.23,116.13) --
	( 59.94,116.09) --
	( 59.66,116.01) --
	( 59.40,115.89) --
	( 59.16,115.72) --
	( 58.95,115.52) --
	( 58.78,115.29) --
	( 58.64,115.03) --
	( 58.55,114.75) --
	( 58.50,114.47) --
	( 58.50,114.32) --
	( 58.50,104.31) --
	( 58.50,104.45) --
	( 58.50,104.16) --
	( 58.55,103.88) --
	( 58.64,103.60) --
	( 58.78,103.34) --
	( 58.95,103.11) --
	( 59.16,102.91) --
	( 59.40,102.74) --
	( 59.66,102.62) --
	( 59.94,102.54) --
	( 60.23,102.50) --
	cycle;
\end{scope}
\begin{scope}
\path[clip] (  8.25,  8.25) rectangle (283.58,139.04);
\definecolor{drawColor}{RGB}{0,0,0}

\node[text=drawColor,anchor=base west,inner sep=0pt, outer sep=0pt, scale=  1.10] at ( 61.82,105.51) {Malay};
\definecolor{fillColor}{RGB}{190,190,190}

\path[fill=fillColor] ( 60.30,126.28) --
	( 84.48,126.28) --
	( 84.41,126.28) --
	( 84.70,126.30) --
	( 84.98,126.35) --
	( 85.26,126.46) --
	( 85.51,126.60) --
	( 85.73,126.79) --
	( 85.93,127.00) --
	( 86.08,127.25) --
	( 86.20,127.52) --
	( 86.26,127.80) --
	( 86.29,128.09) --
	( 86.29,128.09) --
	( 86.29,138.10) --
	( 86.29,138.10) --
	( 86.26,138.39) --
	( 86.20,138.67) --
	( 86.08,138.94) --
	( 85.93,139.19) --
	( 85.73,139.40) --
	( 85.51,139.59) --
	( 85.26,139.73) --
	( 84.98,139.84) --
	( 84.70,139.89) --
	( 84.48,139.91) --
	( 60.30,139.91) --
	( 60.52,139.89) --
	( 60.23,139.91) --
	( 59.94,139.87) --
	( 59.66,139.79) --
	( 59.40,139.67) --
	( 59.16,139.50) --
	( 58.95,139.30) --
	( 58.78,139.07) --
	( 58.64,138.81) --
	( 58.55,138.53) --
	( 58.50,138.25) --
	( 58.50,138.10) --
	( 58.50,128.09) --
	( 58.50,128.23) --
	( 58.50,127.94) --
	( 58.55,127.66) --
	( 58.64,127.38) --
	( 58.78,127.12) --
	( 58.95,126.89) --
	( 59.16,126.69) --
	( 59.40,126.52) --
	( 59.66,126.40) --
	( 59.94,126.32) --
	( 60.23,126.28) --
	cycle;
\end{scope}
\begin{scope}
\path[clip] (  8.25,  8.25) rectangle (283.58,139.04);
\definecolor{drawColor}{RGB}{0,0,0}

\node[text=drawColor,anchor=base west,inner sep=0pt, outer sep=0pt, scale=  1.10] at ( 61.82,129.29) {Iban};
\definecolor{fillColor}{RGB}{190,190,190}

\path[fill=fillColor] ( 60.30, 78.72) --
	(113.06, 78.72) --
	(112.98, 78.72) --
	(113.27, 78.74) --
	(113.56, 78.79) --
	(113.83, 78.90) --
	(114.08, 79.04) --
	(114.31, 79.23) --
	(114.50, 79.44) --
	(114.66, 79.69) --
	(114.77, 79.96) --
	(114.84, 80.24) --
	(114.86, 80.53) --
	(114.86, 80.53) --
	(114.86, 90.54) --
	(114.86, 90.54) --
	(114.84, 90.83) --
	(114.77, 91.11) --
	(114.66, 91.38) --
	(114.50, 91.63) --
	(114.31, 91.84) --
	(114.08, 92.03) --
	(113.83, 92.17) --
	(113.56, 92.28) --
	(113.27, 92.33) --
	(113.06, 92.35) --
	( 60.30, 92.35) --
	( 60.52, 92.33) --
	( 60.23, 92.35) --
	( 59.94, 92.31) --
	( 59.66, 92.23) --
	( 59.40, 92.11) --
	( 59.16, 91.94) --
	( 58.95, 91.74) --
	( 58.78, 91.51) --
	( 58.64, 91.25) --
	( 58.55, 90.97) --
	( 58.50, 90.69) --
	( 58.50, 90.54) --
	( 58.50, 80.53) --
	( 58.50, 80.67) --
	( 58.50, 80.38) --
	( 58.55, 80.10) --
	( 58.64, 79.82) --
	( 58.78, 79.56) --
	( 58.95, 79.33) --
	( 59.16, 79.13) --
	( 59.40, 78.96) --
	( 59.66, 78.84) --
	( 59.94, 78.76) --
	( 60.23, 78.72) --
	cycle;
\end{scope}
\begin{scope}
\path[clip] (  8.25,  8.25) rectangle (283.58,139.04);
\definecolor{drawColor}{RGB}{0,0,0}

\node[text=drawColor,anchor=base west,inner sep=0pt, outer sep=0pt, scale=  1.10] at ( 61.82, 81.73) {Sundanese};
\definecolor{fillColor}{RGB}{107,174,214}

\path[fill=fillColor] ( 60.30, 31.16) --
	(256.62, 31.16) --
	(256.55, 31.16) --
	(256.84, 31.18) --
	(257.12, 31.23) --
	(257.40, 31.34) --
	(257.65, 31.48) --
	(257.87, 31.67) --
	(258.07, 31.88) --
	(258.22, 32.13) --
	(258.33, 32.40) --
	(258.40, 32.68) --
	(258.43, 32.97) --
	(258.43, 32.97) --
	(258.43, 42.98) --
	(258.43, 42.98) --
	(258.40, 43.27) --
	(258.33, 43.55) --
	(258.22, 43.82) --
	(258.07, 44.07) --
	(257.87, 44.28) --
	(257.65, 44.47) --
	(257.40, 44.61) --
	(257.12, 44.72) --
	(256.84, 44.77) --
	(256.62, 44.79) --
	( 60.30, 44.79) --
	( 60.52, 44.77) --
	( 60.23, 44.79) --
	( 59.94, 44.75) --
	( 59.66, 44.67) --
	( 59.40, 44.55) --
	( 59.16, 44.38) --
	( 58.95, 44.18) --
	( 58.78, 43.95) --
	( 58.64, 43.69) --
	( 58.55, 43.41) --
	( 58.50, 43.13) --
	( 58.50, 42.98) --
	( 58.50, 32.97) --
	( 58.50, 33.11) --
	( 58.50, 32.82) --
	( 58.55, 32.54) --
	( 58.64, 32.26) --
	( 58.78, 32.00) --
	( 58.95, 31.77) --
	( 59.16, 31.57) --
	( 59.40, 31.40) --
	( 59.66, 31.28) --
	( 59.94, 31.20) --
	( 60.23, 31.16) --
	cycle;
\end{scope}
\begin{scope}
\path[clip] (  8.25,  8.25) rectangle (283.58,139.04);
\definecolor{drawColor}{RGB}{0,0,0}

\node[text=drawColor,anchor=base west,inner sep=0pt, outer sep=0pt, scale=  1.10] at ( 61.82, 34.17) {Toba Batak {\it dapdap} `a fast-growing tree'};
\definecolor{fillColor}{RGB}{253,141,60}

\path[fill=fillColor] ( 60.30, 54.94) --
	(253.84, 54.94) --
	(253.77, 54.94) --
	(254.06, 54.96) --
	(254.35, 55.01) --
	(254.62, 55.12) --
	(254.87, 55.26) --
	(255.10, 55.45) --
	(255.29, 55.66) --
	(255.44, 55.91) --
	(255.56, 56.18) --
	(255.63, 56.46) --
	(255.65, 56.75) --
	(255.65, 56.75) --
	(255.65, 66.76) --
	(255.65, 66.76) --
	(255.63, 67.05) --
	(255.56, 67.33) --
	(255.44, 67.60) --
	(255.29, 67.85) --
	(255.10, 68.06) --
	(254.87, 68.25) --
	(254.62, 68.39) --
	(254.35, 68.50) --
	(254.06, 68.55) --
	(253.84, 68.57) --
	( 60.30, 68.57) --
	( 60.52, 68.55) --
	( 60.23, 68.57) --
	( 59.94, 68.53) --
	( 59.66, 68.45) --
	( 59.40, 68.33) --
	( 59.16, 68.16) --
	( 58.95, 67.96) --
	( 58.78, 67.73) --
	( 58.64, 67.47) --
	( 58.55, 67.19) --
	( 58.50, 66.91) --
	( 58.50, 66.76) --
	( 58.50, 56.75) --
	( 58.50, 56.89) --
	( 58.50, 56.60) --
	( 58.55, 56.32) --
	( 58.64, 56.04) --
	( 58.78, 55.78) --
	( 58.95, 55.55) --
	( 59.16, 55.35) --
	( 59.40, 55.18) --
	( 59.66, 55.06) --
	( 59.94, 54.98) --
	( 60.23, 54.94) --
	cycle;
\end{scope}
\begin{scope}
\path[clip] (  8.25,  8.25) rectangle (283.58,139.04);
\definecolor{drawColor}{RGB}{0,0,0}

\node[text=drawColor,anchor=base west,inner sep=0pt, outer sep=0pt, scale=  1.10] at ( 61.82, 57.95) {Javanese {\it \textbf{\d{d}a\d{d}}ap} `a species of shade tree'};
\definecolor{fillColor}{RGB}{253,141,60}

\path[fill=fillColor] ( 60.30,  7.38) --
	(239.28,  7.38) --
	(239.21,  7.38) --
	(239.50,  7.40) --
	(239.78,  7.45) --
	(240.05,  7.56) --
	(240.31,  7.70) --
	(240.53,  7.89) --
	(240.72,  8.10) --
	(240.88,  8.35) --
	(240.99,  8.62) --
	(241.06,  8.90) --
	(241.09,  9.19) --
	(241.09,  9.19) --
	(241.09, 19.20) --
	(241.09, 19.20) --
	(241.06, 19.49) --
	(240.99, 19.77) --
	(240.88, 20.04) --
	(240.72, 20.29) --
	(240.53, 20.50) --
	(240.31, 20.69) --
	(240.05, 20.83) --
	(239.78, 20.94) --
	(239.50, 20.99) --
	(239.28, 21.01) --
	( 60.30, 21.01) --
	( 60.52, 20.99) --
	( 60.23, 21.01) --
	( 59.94, 20.97) --
	( 59.66, 20.89) --
	( 59.40, 20.77) --
	( 59.16, 20.60) --
	( 58.95, 20.40) --
	( 58.78, 20.17) --
	( 58.64, 19.91) --
	( 58.55, 19.63) --
	( 58.50, 19.35) --
	( 58.50, 19.20) --
	( 58.50,  9.19) --
	( 58.50,  9.33) --
	( 58.50,  9.04) --
	( 58.55,  8.76) --
	( 58.64,  8.48) --
	( 58.78,  8.22) --
	( 58.95,  7.99) --
	( 59.16,  7.79) --
	( 59.40,  7.62) --
	( 59.66,  7.50) --
	( 59.94,  7.42) --
	( 60.23,  7.38) --
	cycle;
\end{scope}
\begin{scope}
\path[clip] (  8.25,  8.25) rectangle (283.58,139.04);
\definecolor{drawColor}{RGB}{0,0,0}

\node[text=drawColor,anchor=base west,inner sep=0pt, outer sep=0pt, scale=  1.10] at ( 61.82, 10.39) {Sasak {\it \textbf{dad}ap} `a species of shade tree'};
\end{scope}
\end{tikzpicture}

%% file: cognate-CTM.tex
\begin{tikzpicture}[node distance=2cm,->,>=latex,auto,
  every edge/.append style={thick}]
\def\s{1.5}
\definecolor{lightorange1}{HTML}{FED976}
\definecolor{orange1}{HTML}{6BAED6}
\definecolor{blue1}{HTML}{FD8D3C}
\centering
\draw (-5*\s,2*\s) node[circle,minimum size=1.5cm,text width=1.5cm,align=center,draw,fill=gray!40] (absent) {{\sc absent}};
\draw (-3*\s,3*\s) node[circle,minimum size=1.5cm,text width=1.5cm,align=center,draw,fill=orange1] (0) {$-$IC};
\draw (-3*\s,1*\s) node[circle,minimum size=1.5cm,text width=1.5cm,align=center,draw,fill=blue1] (1) {$+$IC};

\draw [dashed,->] (absent) to [bend left=25, style=dashed] node[midway, above left] {$\lambda^{-}$} (0);
\draw [dashed,->] (absent) to [bend right=25, style=dashed] node[midway, below left] {$\lambda^{+}$} (1);
\draw [->] (0) to [bend left=25] node[midway, above left] {$\mu^{-}$} (absent);
\draw [->] (1) to [bend right=25] node[midway, below left] {$\mu^{+}$} (absent);
\draw [->] (0) to [bend left=25] node[midway, above left] {$\rho^{-+}$} (1);
\draw [->] (1) to [bend left=25] node[midway, below right] {$\rho^{+-}$} (0);
\end{tikzpicture}

%% file: cognate-concept-figure.tex
\begin{adjustbox}{max totalsize={.5\linewidth}{.5\linewidth},center}
\begin{minipage}{.6\linewidth}
\begin{adjustbox}{min totalsize={\linewidth}{.5\linewidth},center}


\input{cognate-concept-tree}

\end{adjustbox}
\end{minipage}

\hspace{.25\linewidth}

\begin{minipage}{.5\linewidth}


\begin{adjustbox}{min totalsize={.9\linewidth}{.9\linewidth},center}
\input{cognate-concept-CTM}
\end{adjustbox}

\end{minipage}
\end{adjustbox}

%% file: cognate-concept-tree.tex
\begin{tikzpicture}[x=1pt,y=1pt]
\definecolor{fillColor}{RGB}{255,255,255}
\path[use as bounding box,fill=fillColor,fill opacity=0.00] (0,0) rectangle (180.67,180.67);
\begin{scope}
\path[clip] (  0.00,  0.00) rectangle (180.67,180.67);
\definecolor{drawColor}{RGB}{255,255,255}
\definecolor{fillColor}{RGB}{255,255,255}

\path[draw=drawColor,line width= 0.6pt,line join=round,line cap=round,fill=fillColor] (  0.00,  0.00) rectangle (180.68,180.68);
\end{scope}
\begin{scope}
\path[clip] (  8.25,  8.25) rectangle (175.17,175.17);
\definecolor{drawColor}{RGB}{255,255,255}
\definecolor{fillColor}{RGB}{255,255,255}

\path[draw=drawColor,line width= 0.6pt,line join=round,line cap=round,fill=fillColor] (  8.25,  8.25) rectangle (175.17,175.17);
\definecolor{drawColor}{RGB}{107,174,214}

\path[draw=drawColor,line width= 2.3pt,line join=round,line cap=round] ( 46.95, 53.78) -- ( 77.30, 53.78);

\path[draw=drawColor,line width= 2.3pt,line join=round,line cap=round] ( 46.95, 72.74) -- ( 77.30, 72.74);

\path[draw=drawColor,line width= 2.3pt,line join=round,line cap=round] ( 46.19, 34.81) -- ( 46.95, 34.81);

\path[draw=drawColor,line width= 2.3pt,line join=round,line cap=round] ( 23.43, 15.84) -- ( 77.30, 15.84);

\path[draw=drawColor,line width= 2.3pt,line join=round,line cap=round] ( 38.60,110.68) -- ( 77.30,110.68);
\definecolor{drawColor}{RGB}{253,141,60}

\path[draw=drawColor,line width= 2.3pt,line join=round,line cap=round] ( 46.95,148.62) -- ( 77.30,148.62);
\definecolor{drawColor}{RGB}{107,174,214}

\path[draw=drawColor,line width= 2.3pt,line join=round,line cap=round] ( 54.53,167.59) -- ( 77.30,167.59);
\definecolor{drawColor}{RGB}{253,141,60}

\path[draw=drawColor,line width= 2.3pt,line join=round,line cap=round] ( 46.19,129.65) -- ( 46.95,129.65);
\definecolor{drawColor}{RGB}{190,190,190}

\path[draw=drawColor,line width= 2.3pt,line join=round,line cap=round] ( 38.60, 91.71) -- ( 76.54, 91.71);

\path[] ( 15.84, 70.97) -- ( 15.84, 70.97);
\definecolor{drawColor}{RGB}{107,174,214}

\path[draw=drawColor,line width= 2.3pt,line join=round,line cap=round] ( 15.84, 32.44) -- ( 23.43, 32.44);

\path[draw=drawColor,line width= 2.3pt,line join=round,line cap=round] ( 23.43, 49.03) -- ( 46.19, 49.03);

\path[draw=drawColor,line width= 2.3pt,line join=round,line cap=round] ( 46.19, 63.26) -- ( 46.95, 63.26);

\path[draw=drawColor,line width= 2.3pt,line join=round,line cap=round] ( 15.84,109.50) -- ( 31.01,109.50);

\path[draw=drawColor,line width= 2.3pt,line join=round,line cap=round] ( 31.01,127.28) -- ( 38.60,127.28);

\path[draw=drawColor,line width= 2.3pt,line join=round,line cap=round] ( 38.60,143.88) -- ( 42.39,143.88);
\definecolor{drawColor}{RGB}{253,141,60}

\path[draw=drawColor,line width= 2.3pt,line join=round,line cap=round] ( 42.39,143.88) -- ( 46.19,143.88);

\path[draw=drawColor,line width= 2.3pt,line join=round,line cap=round] ( 46.19,158.10) -- ( 46.95,158.10);

\path[draw=drawColor,line width= 2.3pt,line join=round,line cap=round] ( 46.95,167.59) -- ( 54.53,167.59);
\definecolor{drawColor}{RGB}{107,174,214}

\path[draw=drawColor,line width= 2.3pt,line join=round,line cap=round] ( 31.01, 91.71) -- ( 38.60, 91.71);

\path[draw=drawColor,line width= 2.3pt,line join=round,line cap=round] ( 46.95, 63.26) -- ( 46.95, 53.78);

\path[draw=drawColor,line width= 2.3pt,line join=round,line cap=round] ( 46.95, 63.26) -- ( 46.95, 72.74);

\path[draw=drawColor,line width= 2.3pt,line join=round,line cap=round] ( 46.19, 49.03) -- ( 46.19, 34.81);

\path[draw=drawColor,line width= 2.3pt,line join=round,line cap=round] ( 23.43, 32.44) -- ( 23.43, 15.84);

\path[draw=drawColor,line width= 2.3pt,line join=round,line cap=round] ( 38.60,127.28) -- ( 38.60,110.68);
\definecolor{drawColor}{RGB}{253,141,60}

\path[draw=drawColor,line width= 2.3pt,line join=round,line cap=round] ( 46.95,158.10) -- ( 46.95,148.62);
\definecolor{drawColor}{RGB}{107,174,214}

\path[draw=drawColor,line width= 2.3pt,line join=round,line cap=round] ( 54.53,167.59) -- ( 54.53,167.59);
\definecolor{drawColor}{RGB}{253,141,60}

\path[draw=drawColor,line width= 2.3pt,line join=round,line cap=round] ( 46.19,143.88) -- ( 46.19,129.65);
\definecolor{drawColor}{RGB}{190,190,190}

\path[draw=drawColor,line width= 2.3pt,line join=round,line cap=round] ( 38.60, 91.71) -- ( 38.60, 91.71);

\path[] ( 15.84, 70.97) -- ( 15.84, 70.97);
\definecolor{drawColor}{RGB}{107,174,214}

\path[draw=drawColor,line width= 2.3pt,line join=round,line cap=round] ( 15.84, 70.97) -- ( 15.84, 32.44);

\path[draw=drawColor,line width= 2.3pt,line join=round,line cap=round] ( 23.43, 32.44) -- ( 23.43, 49.03);

\path[draw=drawColor,line width= 2.3pt,line join=round,line cap=round] ( 46.19, 49.03) -- ( 46.19, 63.26);

\path[draw=drawColor,line width= 2.3pt,line join=round,line cap=round] ( 15.84, 70.97) -- ( 15.84,109.50);

\path[draw=drawColor,line width= 2.3pt,line join=round,line cap=round] ( 31.01,109.50) -- ( 31.01,127.28);

\path[draw=drawColor,line width= 2.3pt,line join=round,line cap=round] ( 38.60,127.28) -- ( 38.60,143.88);
\definecolor{drawColor}{RGB}{253,141,60}

\path[draw=drawColor,line width= 2.3pt,line join=round,line cap=round] ( 42.39,143.88) -- ( 42.39,143.88);

\path[draw=drawColor,line width= 2.3pt,line join=round,line cap=round] ( 46.19,143.88) -- ( 46.19,158.10);

\path[draw=drawColor,line width= 2.3pt,line join=round,line cap=round] ( 46.95,158.10) -- ( 46.95,167.59);
\definecolor{drawColor}{RGB}{107,174,214}

\path[draw=drawColor,line width= 2.3pt,line join=round,line cap=round] ( 31.01,109.50) -- ( 31.01, 91.71);
\definecolor{fillColor}{RGB}{107,174,214}

\path[fill=fillColor] ( 79.41, 46.96) --
	(169.69, 46.96) --
	(169.62, 46.96) --
	(169.91, 46.98) --
	(170.19, 47.03) --
	(170.46, 47.14) --
	(170.72, 47.28) --
	(170.94, 47.47) --
	(171.13, 47.68) --
	(171.29, 47.93) --
	(171.40, 48.20) --
	(171.47, 48.48) --
	(171.50, 48.77) --
	(171.50, 48.77) --
	(171.50, 58.78) --
	(171.50, 58.78) --
	(171.47, 59.07) --
	(171.40, 59.35) --
	(171.29, 59.62) --
	(171.13, 59.87) --
	(170.94, 60.08) --
	(170.72, 60.27) --
	(170.46, 60.41) --
	(170.19, 60.52) --
	(169.91, 60.57) --
	(169.69, 60.59) --
	( 79.41, 60.59) --
	( 79.63, 60.57) --
	( 79.34, 60.59) --
	( 79.05, 60.55) --
	( 78.77, 60.47) --
	( 78.51, 60.35) --
	( 78.27, 60.18) --
	( 78.06, 59.98) --
	( 77.88, 59.75) --
	( 77.75, 59.49) --
	( 77.66, 59.21) --
	( 77.61, 58.93) --
	( 77.60, 58.78) --
	( 77.60, 48.77) --
	( 77.61, 48.91) --
	( 77.61, 48.62) --
	( 77.66, 48.34) --
	( 77.75, 48.06) --
	( 77.88, 47.80) --
	( 78.06, 47.57) --
	( 78.27, 47.37) --
	( 78.51, 47.20) --
	( 78.77, 47.08) --
	( 79.05, 47.00) --
	( 79.34, 46.96) --
	cycle;
\end{scope}
\begin{scope}
\path[clip] (  8.25,  8.25) rectangle (175.17,175.17);
\definecolor{drawColor}{RGB}{0,0,0}

\node[text=drawColor,anchor=base west,inner sep=0pt, outer sep=0pt, scale=  1.10] at ( 80.93, 49.97) {Marathi {\IPA pi:\:ne}};
\definecolor{fillColor}{RGB}{107,174,214}

\path[fill=fillColor] ( 79.41, 65.93) --
	(158.01, 65.93) --
	(157.93, 65.93) --
	(158.23, 65.94) --
	(158.51, 66.00) --
	(158.78, 66.11) --
	(159.03, 66.25) --
	(159.26, 66.43) --
	(159.45, 66.65) --
	(159.61, 66.90) --
	(159.72, 67.17) --
	(159.79, 67.45) --
	(159.81, 67.74) --
	(159.81, 67.74) --
	(159.81, 77.75) --
	(159.81, 77.75) --
	(159.79, 78.04) --
	(159.72, 78.32) --
	(159.61, 78.59) --
	(159.45, 78.84) --
	(159.26, 79.05) --
	(159.03, 79.24) --
	(158.78, 79.38) --
	(158.51, 79.49) --
	(158.23, 79.54) --
	(158.01, 79.56) --
	( 79.41, 79.56) --
	( 79.63, 79.54) --
	( 79.34, 79.56) --
	( 79.05, 79.52) --
	( 78.77, 79.44) --
	( 78.51, 79.31) --
	( 78.27, 79.15) --
	( 78.06, 78.95) --
	( 77.88, 78.72) --
	( 77.75, 78.46) --
	( 77.66, 78.18) --
	( 77.61, 77.90) --
	( 77.60, 77.75) --
	( 77.60, 67.74) --
	( 77.61, 67.88) --
	( 77.61, 67.59) --
	( 77.66, 67.31) --
	( 77.75, 67.03) --
	( 77.88, 66.77) --
	( 78.06, 66.54) --
	( 78.27, 66.34) --
	( 78.51, 66.17) --
	( 78.77, 66.05) --
	( 79.05, 65.97) --
	( 79.34, 65.93) --
	cycle;
\end{scope}
\begin{scope}
\path[clip] (  8.25,  8.25) rectangle (175.17,175.17);
\definecolor{drawColor}{RGB}{0,0,0}

\node[text=drawColor,anchor=base west,inner sep=0pt, outer sep=0pt, scale=  1.10] at ( 80.93, 68.94) {Hindi {\IPA pi:na:}};
\definecolor{fillColor}{RGB}{107,174,214}

\path[fill=fillColor] ( 49.06, 27.99) --
	(141.85, 27.99) --
	(141.78, 27.99) --
	(142.07, 28.01) --
	(142.36, 28.06) --
	(142.63, 28.17) --
	(142.88, 28.31) --
	(143.10, 28.50) --
	(143.30, 28.71) --
	(143.45, 28.96) --
	(143.57, 29.23) --
	(143.64, 29.51) --
	(143.66, 29.80) --
	(143.66, 29.80) --
	(143.66, 39.81) --
	(143.66, 39.81) --
	(143.64, 40.10) --
	(143.57, 40.38) --
	(143.45, 40.65) --
	(143.30, 40.90) --
	(143.10, 41.12) --
	(142.88, 41.30) --
	(142.63, 41.44) --
	(142.36, 41.55) --
	(142.07, 41.61) --
	(141.85, 41.62) --
	( 49.06, 41.62) --
	( 49.28, 41.61) --
	( 48.99, 41.62) --
	( 48.70, 41.58) --
	( 48.42, 41.50) --
	( 48.16, 41.38) --
	( 47.92, 41.21) --
	( 47.71, 41.01) --
	( 47.53, 40.78) --
	( 47.40, 40.52) --
	( 47.31, 40.24) --
	( 47.26, 39.96) --
	( 47.25, 39.81) --
	( 47.25, 29.80) --
	( 47.26, 29.95) --
	( 47.26, 29.65) --
	( 47.31, 29.37) --
	( 47.40, 29.09) --
	( 47.53, 28.83) --
	( 47.71, 28.60) --
	( 47.92, 28.40) --
	( 48.16, 28.24) --
	( 48.42, 28.11) --
	( 48.70, 28.03) --
	( 48.99, 27.99) --
	cycle;
\end{scope}
\begin{scope}
\path[clip] (  8.25,  8.25) rectangle (175.17,175.17);
\definecolor{drawColor}{RGB}{0,0,0}

\node[text=drawColor,anchor=base west,inner sep=0pt, outer sep=0pt, scale=  1.10] at ( 50.58, 31.00) {Sanskrit {\IPA pibati}};
\definecolor{fillColor}{RGB}{107,174,214}

\path[fill=fillColor] ( 79.41,  9.02) --
	(171.53,  9.02) --
	(171.46,  9.03) --
	(171.75,  9.04) --
	(172.03,  9.10) --
	(172.30,  9.20) --
	(172.55,  9.34) --
	(172.78,  9.53) --
	(172.97,  9.75) --
	(173.13,  9.99) --
	(173.24, 10.26) --
	(173.31, 10.54) --
	(173.34, 10.83) --
	(173.34, 10.83) --
	(173.34, 20.84) --
	(173.34, 20.84) --
	(173.31, 21.13) --
	(173.24, 21.42) --
	(173.13, 21.68) --
	(172.97, 21.93) --
	(172.78, 22.15) --
	(172.55, 22.33) --
	(172.30, 22.48) --
	(172.03, 22.58) --
	(171.75, 22.64) --
	(171.53, 22.65) --
	( 79.41, 22.65) --
	( 79.63, 22.64) --
	( 79.34, 22.65) --
	( 79.05, 22.61) --
	( 78.77, 22.53) --
	( 78.51, 22.41) --
	( 78.27, 22.24) --
	( 78.06, 22.04) --
	( 77.88, 21.81) --
	( 77.75, 21.55) --
	( 77.66, 21.28) --
	( 77.61, 20.99) --
	( 77.60, 20.84) --
	( 77.60, 10.83) --
	( 77.61, 10.98) --
	( 77.61, 10.69) --
	( 77.66, 10.40) --
	( 77.75, 10.12) --
	( 77.88,  9.87) --
	( 78.06,  9.63) --
	( 78.27,  9.43) --
	( 78.51,  9.27) --
	( 78.77,  9.14) --
	( 79.05,  9.06) --
	( 79.34,  9.03) --
	cycle;
\end{scope}
\begin{scope}
\path[clip] (  8.25,  8.25) rectangle (175.17,175.17);
\definecolor{drawColor}{RGB}{0,0,0}

\node[text=drawColor,anchor=base west,inner sep=0pt, outer sep=0pt, scale=  1.10] at ( 80.93, 12.04) {Ukrainian {\IPA pIte}};
\definecolor{fillColor}{RGB}{107,174,214}

\path[fill=fillColor] ( 79.41,103.87) --
	(156.35,103.87) --
	(156.28,103.87) --
	(156.57,103.88) --
	(156.85,103.94) --
	(157.13,104.04) --
	(157.38,104.19) --
	(157.60,104.37) --
	(157.80,104.59) --
	(157.95,104.84) --
	(158.07,105.10) --
	(158.14,105.39) --
	(158.16,105.67) --
	(158.16,105.67) --
	(158.16,115.69) --
	(158.16,115.69) --
	(158.14,115.98) --
	(158.07,116.26) --
	(157.95,116.53) --
	(157.80,116.77) --
	(157.60,116.99) --
	(157.38,117.17) --
	(157.13,117.32) --
	(156.85,117.42) --
	(156.57,117.48) --
	(156.35,117.49) --
	( 79.41,117.49) --
	( 79.63,117.48) --
	( 79.34,117.49) --
	( 79.05,117.46) --
	( 78.77,117.38) --
	( 78.51,117.25) --
	( 78.27,117.09) --
	( 78.06,116.89) --
	( 77.88,116.65) --
	( 77.75,116.40) --
	( 77.66,116.12) --
	( 77.61,115.83) --
	( 77.60,115.69) --
	( 77.60,105.67) --
	( 77.61,105.82) --
	( 77.61,105.53) --
	( 77.66,105.24) --
	( 77.75,104.97) --
	( 77.88,104.71) --
	( 78.06,104.48) --
	( 78.27,104.28) --
	( 78.51,104.11) --
	( 78.77,103.99) --
	( 79.05,103.90) --
	( 79.34,103.87) --
	cycle;
\end{scope}
\begin{scope}
\path[clip] (  8.25,  8.25) rectangle (175.17,175.17);
\definecolor{drawColor}{RGB}{0,0,0}

\node[text=drawColor,anchor=base west,inner sep=0pt, outer sep=0pt, scale=  1.10] at ( 80.93,106.88) {Breton {\IPA evA}};
\definecolor{fillColor}{RGB}{107,174,214}

\path[fill=fillColor] ( 79.41,160.77) --
	(180.67,160.77) --
	(180.67,174.40) --
	( 79.41,174.40) --
	( 79.63,174.39) --
	( 79.34,174.40) --
	( 79.05,174.36) --
	( 78.77,174.28) --
	( 78.51,174.16) --
	( 78.27,173.99) --
	( 78.06,173.79) --
	( 77.88,173.56) --
	( 77.75,173.30) --
	( 77.66,173.03) --
	( 77.61,172.74) --
	( 77.60,172.59) --
	( 77.60,162.58) --
	( 77.61,162.73) --
	( 77.61,162.44) --
	( 77.66,162.15) --
	( 77.75,161.87) --
	( 77.88,161.62) --
	( 78.06,161.38) --
	( 78.27,161.18) --
	( 78.51,161.02) --
	( 78.77,160.89) --
	( 79.05,160.81) --
	( 79.34,160.78) --
	cycle;
\end{scope}
\begin{scope}
\path[clip] (  8.25,  8.25) rectangle (175.17,175.17);
\definecolor{drawColor}{RGB}{0,0,0}

\node[text=drawColor,anchor=base west,inner sep=0pt, outer sep=0pt, scale=  1.10] at ( 80.93,163.79) {French {\IPA bwa\textscr}};
\definecolor{fillColor}{RGB}{253,141,60}

\path[fill=fillColor] ( 79.41,141.81) --
	(180.67,141.81) --
	(180.67,155.43) --
	( 79.41,155.43) --
	( 79.63,155.42) --
	( 79.34,155.43) --
	( 79.05,155.40) --
	( 78.77,155.31) --
	( 78.51,155.19) --
	( 78.27,155.02) --
	( 78.06,154.82) --
	( 77.88,154.59) --
	( 77.75,154.33) --
	( 77.66,154.06) --
	( 77.61,153.77) --
	( 77.60,153.62) --
	( 77.60,143.61) --
	( 77.61,143.76) --
	( 77.61,143.47) --
	( 77.66,143.18) --
	( 77.75,142.90) --
	( 77.88,142.65) --
	( 78.06,142.41) --
	( 78.27,142.21) --
	( 78.51,142.05) --
	( 78.77,141.92) --
	( 79.05,141.84) --
	( 79.34,141.81) --
	cycle;
\end{scope}
\begin{scope}
\path[clip] (  8.25,  8.25) rectangle (175.17,175.17);
\definecolor{drawColor}{RGB}{0,0,0}

\node[text=drawColor,anchor=base west,inner sep=0pt, outer sep=0pt, scale=  1.10] at ( 80.93,144.82) {Portuguese {\IPA {\bf b\textbari b}e\textfishhookr}};
\definecolor{fillColor}{RGB}{253,141,60}

\path[fill=fillColor] ( 49.06,122.84) --
	(129.68,122.84) --
	(129.61,122.84) --
	(129.90,122.85) --
	(130.18,122.91) --
	(130.46,123.01) --
	(130.71,123.16) --
	(130.93,123.34) --
	(131.13,123.56) --
	(131.28,123.80) --
	(131.39,124.07) --
	(131.46,124.35) --
	(131.49,124.64) --
	(131.49,124.64) --
	(131.49,134.66) --
	(131.49,134.66) --
	(131.46,134.95) --
	(131.39,135.23) --
	(131.28,135.50) --
	(131.13,135.74) --
	(130.93,135.96) --
	(130.71,136.14) --
	(130.46,136.29) --
	(130.18,136.39) --
	(129.90,136.45) --
	(129.68,136.46) --
	( 49.06,136.46) --
	( 49.28,136.45) --
	( 48.99,136.46) --
	( 48.70,136.43) --
	( 48.42,136.35) --
	( 48.16,136.22) --
	( 47.92,136.06) --
	( 47.71,135.85) --
	( 47.53,135.62) --
	( 47.40,135.36) --
	( 47.31,135.09) --
	( 47.26,134.80) --
	( 47.25,134.66) --
	( 47.25,124.64) --
	( 47.26,124.79) --
	( 47.26,124.50) --
	( 47.31,124.21) --
	( 47.40,123.94) --
	( 47.53,123.68) --
	( 47.71,123.45) --
	( 47.92,123.24) --
	( 48.16,123.08) --
	( 48.42,122.95) --
	( 48.70,122.87) --
	( 48.99,122.84) --
	cycle;
\end{scope}
\begin{scope}
\path[clip] (  8.25,  8.25) rectangle (175.17,175.17);
\definecolor{drawColor}{RGB}{0,0,0}

\node[text=drawColor,anchor=base west,inner sep=0pt, outer sep=0pt, scale=  1.10] at ( 50.58,125.85) {Latin {\IPA {\bf bib}ere}};
\definecolor{fillColor}{RGB}{190,190,190}

\path[fill=fillColor] ( 78.65, 84.90) --
	(167.92, 84.90) --
	(167.85, 84.90) --
	(168.14, 84.91) --
	(168.42, 84.97) --
	(168.69, 85.07) --
	(168.94, 85.22) --
	(169.17, 85.40) --
	(169.36, 85.62) --
	(169.52, 85.87) --
	(169.63, 86.13) --
	(169.70, 86.42) --
	(169.73, 86.71) --
	(169.73, 86.71) --
	(169.73, 96.72) --
	(169.73, 96.72) --
	(169.70, 97.01) --
	(169.63, 97.29) --
	(169.52, 97.56) --
	(169.36, 97.80) --
	(169.17, 98.02) --
	(168.94, 98.21) --
	(168.69, 98.35) --
	(168.42, 98.45) --
	(168.14, 98.51) --
	(167.92, 98.53) --
	( 78.65, 98.53) --
	( 78.87, 98.51) --
	( 78.58, 98.52) --
	( 78.29, 98.49) --
	( 78.01, 98.41) --
	( 77.75, 98.28) --
	( 77.51, 98.12) --
	( 77.30, 97.92) --
	( 77.12, 97.68) --
	( 76.99, 97.43) --
	( 76.90, 97.15) --
	( 76.85, 96.86) --
	( 76.84, 96.72) --
	( 76.84, 86.71) --
	( 76.85, 86.85) --
	( 76.85, 86.56) --
	( 76.90, 86.27) --
	( 76.99, 86.00) --
	( 77.12, 85.74) --
	( 77.30, 85.51) --
	( 77.51, 85.31) --
	( 77.75, 85.14) --
	( 78.01, 85.02) --
	( 78.29, 84.94) --
	( 78.58, 84.90) --
	cycle;
\end{scope}
\begin{scope}
\path[clip] (  8.25,  8.25) rectangle (175.17,175.17);
\definecolor{drawColor}{RGB}{0,0,0}

\node[text=drawColor,anchor=base west,inner sep=0pt, outer sep=0pt, scale=  1.10] at ( 80.17, 87.91) {English {\IPA drINk}};
\end{scope}
\end{tikzpicture}

%% file: cognate-concept-CTM.tex
\begin{tikzpicture}[node distance=2cm,->,>=latex,auto,
  every edge/.append style={thick}]
\def\s{1.5}
\definecolor{lightorange1}{HTML}{6BAED6}
\definecolor{orange1}{HTML}{FD8D3C}
\definecolor{blue1}{HTML}{C6DBEF}
\centering
\draw (-5*\s,2*\s) node[circle,minimum size=1.5cm,text width=1.5cm,align=center,draw,fill=lightgray] (drink0) {{\sc absent}};
\draw (-3*\s,3*\s) node[circle,minimum size=1.5cm,text width=1.5cm,align=center,draw,fill=lightorange1] (peH0) {*peh$_3$- $-${\sc IC}};
\draw (-3*\s,1*\s) node[circle,minimum size=1.5cm,text width=1.5cm,align=center,draw,fill=orange1] (peH1) {*peh$_3$- $+${\sc IC}};
\draw [->] (drink0) to [bend left=25] node[midway, above left] {$\lambda^{-}$} (peH0);
\draw [->] (drink0) to [bend right=25] node[midway, below left] {$\lambda^{+}$} (peH1);
\draw [->] (peH0) to [bend left=25] node[midway, above left] {$\mu^{-}$} (drink0);
\draw [->] (peH1) to [bend right=25] node[midway, below left] {$\mu^{+}$} (drink0);
\draw [->] (peH0) to [bend left=25] node[midway, above left] {$\rho^{-+}$} (peH1);
\draw [->] (peH1) to [bend left=25] node[midway, below right] {$\rho^{+-}$} (peH0);

\end{tikzpicture}

%% file: results.tex
\begin{table}[t]
\centering

\small{Cognate class traits}

\footnotesize{
    \begin{tabular}{|p{.2\linewidth}|p{.2\linewidth}|p{.5\linewidth}|}
    \hline
        Rate 1 & Rate 2 & Question addressed \\
        \hline
        \hline
        $\lambda^{-}_0$ \newline (birth rate, $-$IC) & $\lambda^{+}_0$ \newline (birth rate, $+$IC) & Do word forms without IC arise more frequently than forms with IC? \newline (Yes, more frequently than at chance)\\
        \hline
        $\rho^{+-}_0$ \newline (mut.\ rate, $+$IC$\rightarrow-$IC) & $\rho^{-+}_0$ \newline (mut.\ rate, $+$IC$\rightarrow-$IC) & Are $+$IC$\rightarrow-$IC changes more frequent than $-$IC$\rightarrow+$IC changes? \newline (Yes, but not always more frequently than at chance)\\
        \hline
        $\mu^{+}_0$ \newline (loss rate, $+$IC) & $\mu^{-}_0$ \newline (loss rate, $-$IC) & Are forms with $+$IC more likely to die out than forms with $-$IC? \newline (No)\\
        \hline
    \end{tabular}
    }

\small{Cognate-concept traits}

\footnotesize{
    \begin{tabular}{|p{.2\linewidth}|p{.2\linewidth}|p{.5\linewidth}|}
    \hline
        Rate 1 & Rate 2 & Question addressed \\
        \hline
        \hline
        $\lambda^{-}_0$ \newline (birth rate, $-$IC) & $\lambda^{+}_0$ \newline (birth rate, $+$IC) & Do word forms without IC enter the basic vocabulary more frequently than forms with IC? \newline (Yes, but not more frequently than chance in 4/5 families)\\
        \hline
        $\rho^{+-}_0$ \newline (mut.\ rate, $+$IC$\rightarrow-$IC) & $\rho^{-+}_0$ \newline (mut.\ rate, $+$IC$\rightarrow-$IC) & Are $+$IC$\rightarrow-$IC changes more frequent than $-$IC$\rightarrow+$IC changes in basic vocabulary items? \newline (Greater than chance in only 1/5 families)\\
        \hline
        $\mu^{+}_0$ \newline (loss rate, $+$IC) & $\mu^{-}_0$ \newline (loss rate, $-$IC) & Are forms with $+$IC phased out of basic meaning functions more often than forms with $-$IC? \newline (Yes.)\\
        \hline
    \end{tabular}
    }

    \caption{Interpretation of parameters used in analyses of cognate class and cognate-concept traits, along with the research questions they are used to address as well as answers. 
    Subscript zeros indicate that parameters represent log mean rates around which rates for individual traits are log-normally distributed (with the exception of $\lambda^{\pm}_0$ for cognate class traits; see text).
    Each hypothesis is assessed by computing the ratio between rates 1 and 2 after exponentiating them. 
    }
    \label{tab:params}
\end{table}

\subsection{Cognate class traits}

Bayesian phylogenetic models were used to disentangle the mechanisms that shape the evolutionary trajectories of individual cognate classes (e.g., forms descending from Proto-Malayo-Polynesian {*dapdap}) in three families (Austronesian, Semitic, and Uralic). 
Over the course of a language family's phylogenetic history, 
ancestral word forms are born, undergo processes of word form mutation and differentiation (as the speech varieties in which they exist diversify phylogenetically), and die out on different phylogenetic lineages. 
Analyses of the evolution of morpheme-internal identical consonants within cognate class traits in three language families 
were carried out using a hierarchical phylogenetic model containing six parameters of interest (schematized in Figure \ref{fig:ctm} and further defined in Table \ref{tab:params}): $\lambda^{-}_0$, the log birth rate of forms without identical consonants; 
$\lambda^{+}_0$, the log birth rate of forms with identical consonants; 
$\rho^{-+}_0$, the log mean rate at which sequences of identical consonants arise within forms; 
$\rho^{+-}_0$, the log mean rate at which sequences of identical consonants are lost within forms; 
$\mu^{-}_0$, the log mean loss rate of forms without identical consonants; 
and $\mu^{+}_0$, the log mean loss rate of forms with identical consonants. 
The hierarchical model used allows parameters to vary at the level of individual cognate classes, 
which undergo change according to evolutionary rates that are log-normally distributed around the mean parameters 
$\rho^{-+}_0$, $\rho^{+-}_0$, $\mu^{-}_0$, $\mu^{+}_0$, or in the case of birth rates, according to which all cognate classes arise and which are shared across all cognate classes, set to $\exp\left( \lambda^{-}_0 \right)$ and $\exp\left( \lambda^{+}_0 \right)$. 
Parameters that vary at the level of individual cognate classes are analogous to random effects in mixed-effects regression models, in that they account for individual cognate-level idiosyncrasies, while the mean parameters listed above are comparable to fixed effects, as they capture global trends in the evolutionary system. 
Pairwise comparisons between parameters allow us to assess whether forms with and without identical consonants are born at different rates ($\lambda^{+}_0$ vs.\ $\lambda^{-}_0$), whether identical consonants are gained and lost within forms at different rates ($\rho^{-+}_0$ vs.\ $\rho^{+-}_0$), and whether forms with and without identical consonants are lost at different rates ($\mu^{+}_0$ vs.\ $\mu^{-}_0$). 
Strengths of differences in rates were quantified by taking the ratio of the two mean rates in question, i.e., by inspecting the posterior distributions of the quantities $\exp\left( \lambda^{-}_0 - \lambda^{+}_0 \right)$, $\exp\left( \rho^{+-}_0 - \rho^{-+}_0 \right)$, and $\exp\left( \mu^{+}_0 - \mu^{-}_0 \right)$. 
Evidence for a difference is taken to be decisive if the 95\% highest density interval 
of ratios does not contain values representing the null hypothesis \cite{kruschke2021bayesian}. 
A standard null value is 1: ratios greater than 1 indicate that one change type is more frequent than another. 
However, in some cases, skewed distributions are expected even under null models of language generation \cite{moscoso2013missing}. 
For this reason, posterior ratios are also compared to quantities representing baseline asymmetries in frequencies of change types that would be expected under neutral processes of language evolution. 

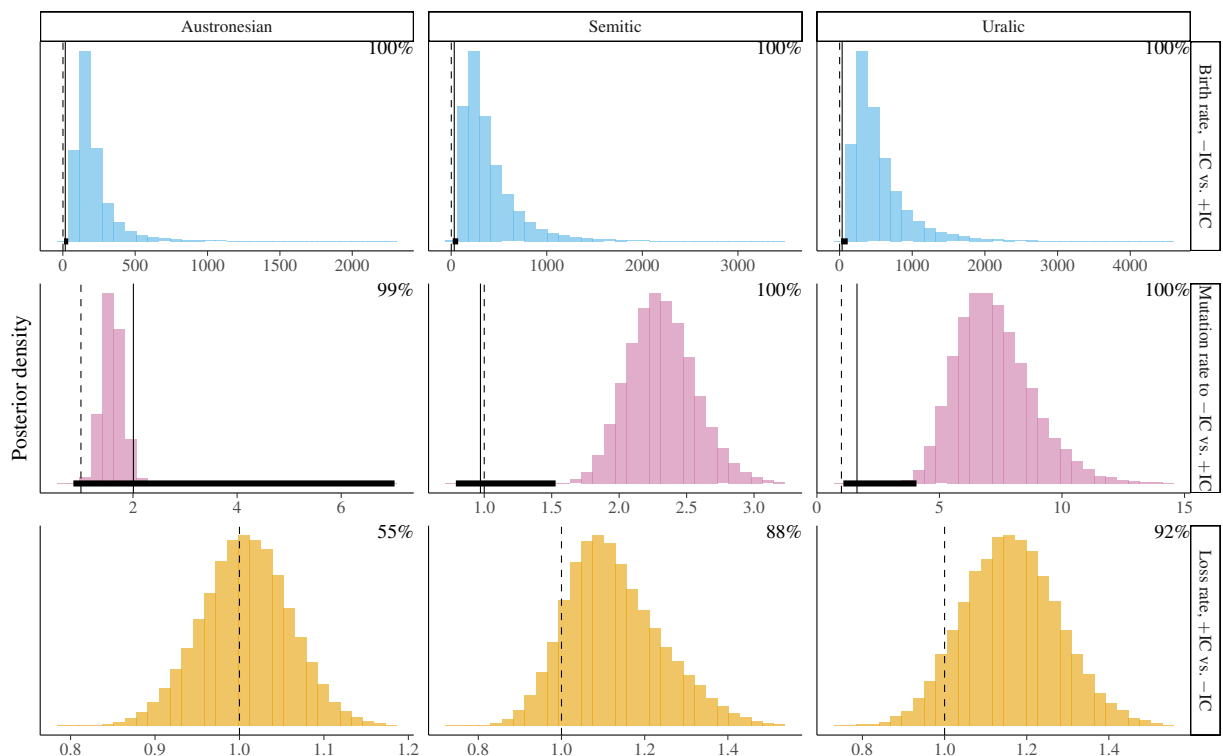
\begin{figure}[t]
    \centering
    \begin{adjustbox}{max totalsize={\linewidth}{\linewidth},center}

        \small{
        \input{cognate_posterior_rates}

        }
        
        \end{adjustbox}
    \caption{Histograms from analyses of cognate traits displaying posterior distributions of ratios of 
    parameters of interest for different families: birth rate of words with value $-$IC (no identical consonants) vs.\ $+$IC (with identical consonants; top), 
    rate of $+$IC $\rightarrow$ $-$IC vs.\ $-$IC $\rightarrow$ $+$IC change (middle), 
    and loss rate of words with $+$IC vs.\ $-$IC (bottom). Histograms are annotated with percentages of samples for which ratios are greater than 1 (given by vertical dashed lines). Solid black vertical lines in upper two rows represent median baseline quantities;  horizontal lines represent ranges of baseline quantities.}
    \label{fig:cognate-posteriors}
\end{figure}

Figure \ref{fig:cognate-posteriors} shows posterior distributions of ratios of interest. Distributions are annotated with the percentage of posterior samples for which the ratio is greater than one (represented by dashed lines). 
Distributions of ratios pertaining to birth rates and mutation rates are also annotated with values representing ranges of ratios (and median values thereof) that would be expected under neutral models of language change. 
These quantities are estimated from data from each family under analysis, 
assuming 
that distributions of features found in contemporary languages are representative of those encountered during the history of the language family to which they belong \cite{Lass1993}. 
Under a neutral 
process 
in which words are generated by randomly sampling segments with uniform probabilities, the ratio of words born without versus with sequences of identical consonants is no greater than the number of consonants in a language's segmental inventory, minus one. This quantity is provided for languages in each family for which such data are available. 
Baseline values for ratios between mutations that remove versus introduce sequences of identical consonants 
are estimated by simulating the effects of neutral models of sound change \cite{beguvs2020estimating,ceolin2020neutral-key} using word lists of languages in the families under study. 
For loss rates, 
a baseline value of 1 is sufficient for the purpose of interpreting posterior ratios.



Across all three families, there is decisive evidence that forms without identical consonants are born more frequently than those with identical consonants (median: $171.49$, 95\% HDI: [$51.67$, $526.09$]; $303.07$, [$53.17$, $1017.29$]; and $428.73$, [$89.93$, $1365.98$] times more frequently in Austronesian, Semitic, and Uralic, respectively). 
Additionally, there is decisive evidence that these ratios are greater than would be expected under a chance baseline based on sizes of segmental inventories, as posterior HDIs 
are consistently greater in value than 
ranges of baselines expected under a neutral process of word generation (median: $17$, total range: [$8$, $37$]; $32$, [$16$, $73$]; and $35$, [$21$, $112$] times more frequently in Austronesian, Semitic, and Uralic, respectively). 
Mutational changes to forms that remove sequences of identical consonants are decisively more frequent than mutations that introduce them, although the ratios between transition rates pertaining to these changes are far lower than asymmetries in birth rates of forms with and without identical consonants ($1.6$, [$1.23$, $1.99$]; $2.30$, [$1.85$, $2.82$]; and $7.09$, [$4.44$, $10.45$] in Austronesian, Semitic, and Uralic, respectively). 
95\% HDIs overlap with ratios expected under neutral models of sound change in Austronesian but not Semitic and Uralic ($2.00$, [$0.85$, $7.02$]; $0.97$, [$0.78$, $1.52$]; and $1.63$, [$1.08$, $4.06$] more frequently in Austronesian, Semitic, and Uralic, respectively), indicating that these ratios exceed what is expected at chance levels only in the latter two families. 
Posterior distributions do not support the idea that forms with sequences of identical consonants die out more frequently than those without them ($1.01$, [$0.9$, $1.11$]; $1.11$, [$0.92$, $1.37$]; and $1.16$, [$0.94$, $1.39$] in Austronesian, Semitic, and Uralic, respectively. 



These results indicate that asymmetries in birth rates of words play a major and consistent 
role in the under-representation of sequences of identical consonants in word forms, and to a weaker extent processes that mutate word forms, though this latter effect is not found in all families studied when interpreted according to a principled, conservative baseline. 
Crucially, however, word forms containing such sequences are no more likely to fall entirely out of use than those without: they 
exhibit 
as much longevity as their counterparts that do not contain identical consonants, 
though it is not clear from these results whether 
they survive in more marginal functions and restricted distributions. 

\subsection{Cognate-concept traits}
A related set of phylogenetic models were used to analyze the evolution of morpheme-internal sequences of identical consonants within cognate-concept traits in five language families (Dravidian, Indo-European, Sino-Tibetan, Turkic, and Uto-Aztecan). 
These analyses shed light on the conditions under which cognate word forms enter and fall out of use in 
basic meaning functions, 
and the nature of the processes affecting word forms during the time in which they occupy 
such roles. 
Analyses focused on cognate-concept traits pertaining to one hundred concepts representing basic vocabulary items, chosen to maximize comparability of results across families \cite{Swadesh1955}. 
Parameters of interest have the similar interpretations as for the models described in the previous section (see Figure \ref{fig:ctm}, Table \ref{tab:params}). 
As above, posterior parameter values were compared to assess whether 
word forms without identical consonants enter basic vocabulary meaning functions more frequently than those without 
($\lambda^{-}_0$ vs.\ $\lambda^{+}_0$), 
whether identical consonants are lost within forms used in the basic vocabulary more frequently than they are gained ($\rho^{+-}_0$ vs.\ $\rho^{-+}_0$), 
and whether forms containing identical consonants are removed from the basic vocabulary more frequently than those without 
($\mu^{+}_0$ vs.\ $\mu^{-}_0$). 
The baselines against which ratios for birth and mutation rates are compared differ from those employed for cognate class traits. 
Ratios of birth rates (i.e., between the rates at which forms without and with identical consonants enter languages' basic vocabulary) are compared to ratios between numbers of forms without versus containing identical consonants in contemporary languages' basic and non-basic vocabularies; 
this comparison tells us whether forms with identical consonants enter the basic vocabulary at a rate lower than would be expected from a neutral process in which  
basic vocabulary items are sampled randomly from the lexicon of a language. 
Ratios between mutation rates are compared to baselines generated via simulations of neutral sound change, as for cognate class traits, but restricted to forms expressing the one hundred concepts under analysis. 
As with cognate class traits, ratios between rates at which forms with and without identical consonants are removed from the basic vocabulary 
do not require interpretation against a baseline other than the standard null value of 1. 

\begin{figure}
    \centering
    \begin{adjustbox}{max totalsize={\linewidth}{\linewidth},center}

        \small{
        \input{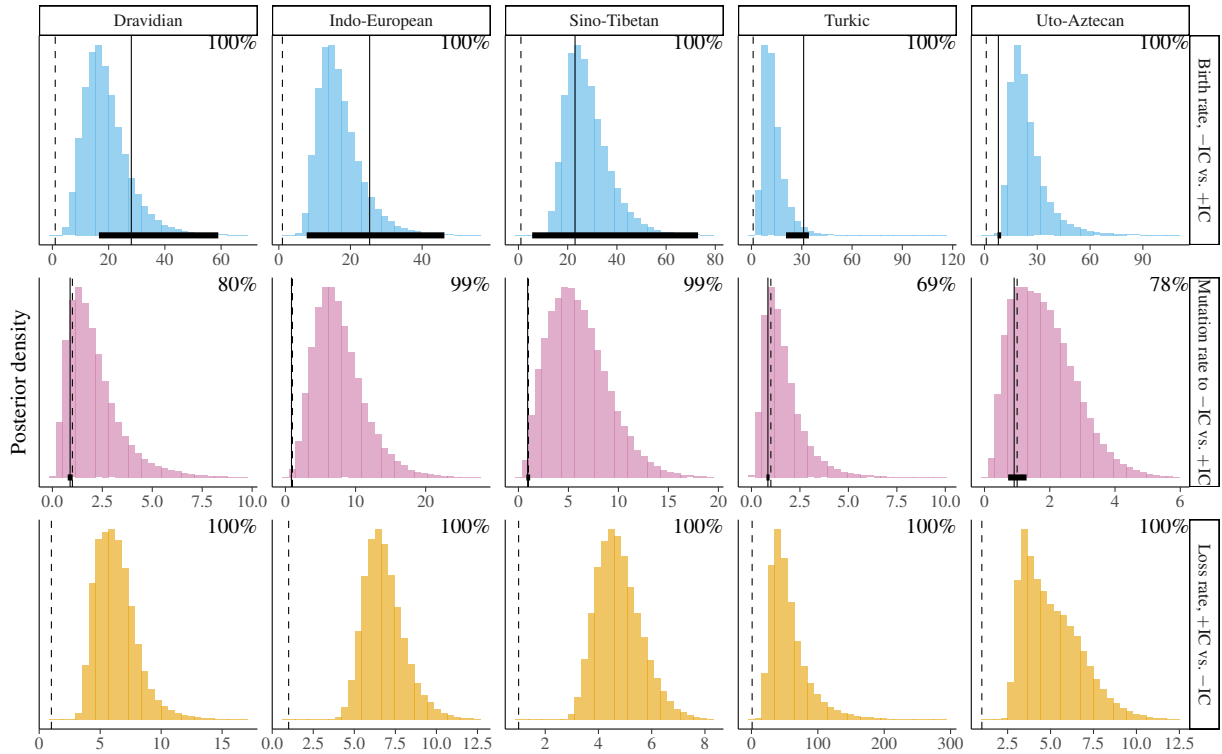}

        }
        \end{adjustbox}

    \caption{Histograms from analyses of cognate-concept traits displaying posterior distributions of ratios of 
    parameters of interest for different families: birth rate of cognate-concept traits with $-$IC vs.\ $+$IC (top), 
    rate of $+$IC $\rightarrow$ $-$IC vs.\ $-$IC $\rightarrow$ $+$IC change (middle) within cognate-concept traits, 
    and loss rate of cognate-concept traits with $+$IC vs.\ $-$IC (bottom). Histograms are annotated with percentages of samples for which ratios are greater than 1 (given by vertical dashed lines). Solid black vertical lines in upper two rows represent median baseline quantities; horizontal lines represent ranges of baseline quantities.}
    \label{fig:cognate-concept-posteriors}
\end{figure}

Figure \ref{fig:cognate-concept-posteriors} shows posterior distributions of ratios of interest. 
Distributions are annotated as in Figure \ref{fig:cognate-posteriors}. 
All families show decisive evidence that forms without identical consonants enter the basic vocabulary more frequently than forms with identical consonants 
(\ExecuteMetaData[cognate-concept-CIs.tex]{birthRate}); 
however, these distributions overlap with ranges of ratios expected under a random sampling process from the lexicon in all families (Dravidian: $28$, [$16.5$, $59$]; Indo-European: $25.3$, [$7.78$, $46.2$]; Sino-Tibetan: $23.0$, [$5.56$, $73$]; Turkic: $31.1$, [$20.6$, $34.2$]; Uto-Aztecan: $7.89$, [$7.42$, $9.46$]) except for Uto-Aztecan, where usable digitized word lists comprising basic and non-basic vocabulary items were available for only three languages.

Indo-European is the only family exhibiting decisive evidence that mutational processes remove sequences of identical consonants from basic vocabulary items more frequently than they introduce them 
(\ExecuteMetaData[cognate-concept-CIs.tex]{gainRate}); 
for Sino-Tibetan, the less conservative 89\% HDI ([$1.24$, $10.18$]) does not overlap with one. 
Indo-European posterior ratios do not overlap with ranges that would be expected under neutral processes of sound change affecting the basic vocabulary (Dravidian: $0.88$, [$0.76$, $1.01$]; Indo-European: $0.86$, [$0.77$, $0.92$]; Sino-Tibetan: $0.92$, [$0.78$, $1.16$]; Turkic: $0.84$, [$0.77$, $0.93$]; Uto-Aztecan: $0.91$, [$0.72$, $1.28$]). 

All families show decisive support for the idea that cognate-concept traits are lost more frequently when the form expressing the concept in question contains identical consonants than when it does not (\ExecuteMetaData[cognate-concept-CIs.tex]{deathRate}). 
This indicates that while word forms with identical consonants do not exhibit less overall longevity than word forms without identical consonants, they are phased out of 
basic meaning functions 
more frequently than those without.

\begin{figure}
    \centering



   \includegraphics[width=6.5in]{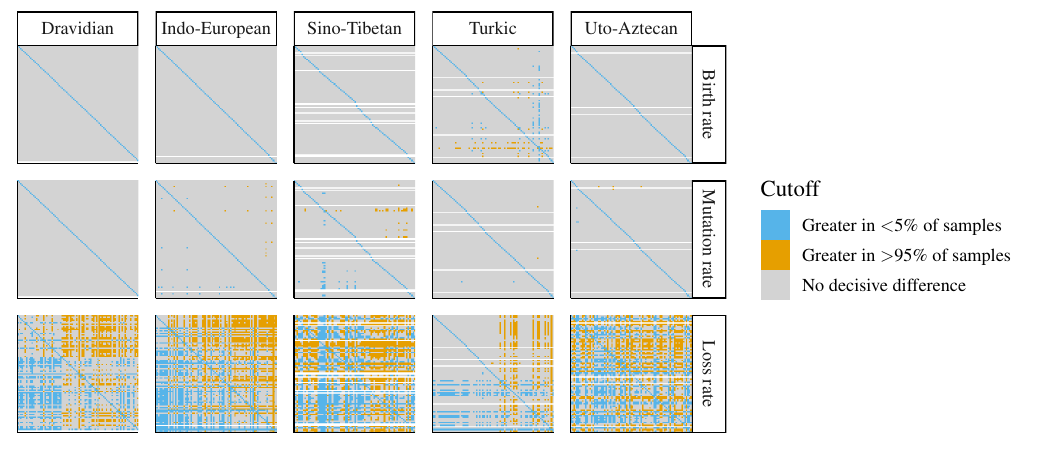}
    
        \caption{Heatmaps displaying pairwise contrasts between concepts, with concepts (represented by heatmap rows and columns, with labels removed for clarity) organized according to a ranking of basicness and stability. In the upper (right) triangle of each heatmap, orange cells indicate contrasts where a more salient concept exhibits a higher asymmetry than a less salient one, blue cells indicate contrasts where a less salient concept exhibits a higher asymmetry than a more salient one, and gray cells indicate no decisive difference.}
    \label{fig:heatmap}
\end{figure}

The rates reported above characterize the dynamics of lexical replacement within the basic vocabulary 
as a whole. 
Variation among rates was inspected at the concept level in order to investigate the extent to which relative strengths of ratios between rates vary across concepts. 
Pairwise comparisons of ratio strength between concepts were carried out 
by 
computing the percentage of samples for which a ratio between rates (birth, mutation, and loss) was greater in one concept than another, with evidence for a contrast taken to be decisive 
for percentages of 95\% or more 
\cite{gelman2012we}. 
Few comparisons involving ratios between birth rates exhibit decisive evidence for a difference 
(Dravidian: 0 out of 4278 pairwise comparisons; Indo-European: 0/4560; Sino-Tibetan: 1/3403; Turkic: 83/4005; Uto-Aztecan: 0/4186), 
along with mutation rates 
(Dravidian: 0/4278; Indo-European: 30/4560; Sino-Tibetan: 48/3403; Turkic: 2/4005; Uto-Aztecan: 6/4186). 
Ratios between loss rates ($+$IC vs.\ $-$IC) exhibit a higher number of decisive contrasts 
(Dravidian: 1469/4278; Indo-European: 2122/4560; Sino-Tibetan: 2038/3403; Turkic: 666/4005; Uto-Aztecan: 2132/4186), 
indicating that while on the whole, forms with sequences of identical consonants are phased out of the basic vocabulary at a higher rate than forms without, 
the strength of this tendency differs considerably across concepts. 
The heatmaps in Figure \ref{fig:heatmap} display pairwise contrasts between concepts. 
Concepts are organized according to 
a ranking of basicness and stability \cite{dellert2018new}, 
with lower values indicating more salient and usually more frequently used \cite{Pagel2007,calude2014frequency} concepts and higher values indicating more marginal ones. 
In the upper (right) triangle of each heatmap, orange cells indicate contrasts where a more salient concept exhibits a higher ratio between rates than a less salient one, blue cells indicate contrasts where a less salient concept exhibits a higher ratio between rates than a more salient one, and gray cells indicate that there is no decisive difference between concepts. 
With the exception of Turkic, the majority of contrasts concerning asymmetries in loss rates exhibit a decisively higher asymmetry for the more salient concept, indicating that rates at which forms with identical consonants are phased out of the basic vocabulary relative to forms without identical consonants is weaker for less salient concepts. 

In sum, analyses of the evolution of concept-cognate traits show that while forms without sequences of identical consonants enter the basic vocabulary more frequently than forms containing such sequences, these ratios are comparable (in all but one family) to what would be expected from sampling at random from the lexicons of contemporary languages in each family. 
There is not clear evidence across families that mutation rates are more likely to remove sequences of identical consonants than introduce them. 
Loss rates for cognate-concept traits consistently display rate asymmetries in favor of forms without identical consonants. 
At the same time, for loss rates, the strength of these asymmetries is weaker for less salient concepts. Extrapolating these dynamics beyond the basic vocabulary, forms with identical consonants can be expected to survive in concept slots that are infrequently used and marginal for roughly as long as forms without. 



%% file: cognate_posterior_rates.tex
\begin{tikzpicture}[x=1pt,y=1pt]
\definecolor{fillColor}{RGB}{255,255,255}
\path[use as bounding box,fill=fillColor,fill opacity=0.00] (0,0) rectangle (682.95,439.04);
\begin{scope}
\path[clip] (  0.00,  0.00) rectangle (682.95,439.04);
\definecolor{drawColor}{RGB}{255,255,255}
\definecolor{fillColor}{RGB}{255,255,255}

\path[draw=drawColor,line width= 0.6pt,line join=round,line cap=round,fill=fillColor] (  0.00,  0.00) rectangle (682.95,439.04);
\end{scope}
\begin{scope}
\path[clip] ( 20.71,300.36) rectangle (228.60,416.97);
\definecolor{fillColor}{RGB}{255,255,255}

\path[fill=fillColor] ( 20.71,300.36) rectangle (228.60,416.97);
\definecolor{fillColor}{RGB}{86,180,233}

\path[fill=fillColor,fill opacity=0.60] ( 30.16,305.66) rectangle ( 36.46,305.66);

\path[fill=fillColor,fill opacity=0.60] ( 36.46,305.66) rectangle ( 42.76,356.58);

\path[fill=fillColor,fill opacity=0.60] ( 42.76,305.66) rectangle ( 49.06,411.67);

\path[fill=fillColor,fill opacity=0.60] ( 49.06,305.66) rectangle ( 55.36,357.50);

\path[fill=fillColor,fill opacity=0.60] ( 55.36,305.66) rectangle ( 61.66,327.04);

\path[fill=fillColor,fill opacity=0.60] ( 61.66,305.66) rectangle ( 67.96,316.33);

\path[fill=fillColor,fill opacity=0.60] ( 67.96,305.66) rectangle ( 74.26,311.29);

\path[fill=fillColor,fill opacity=0.60] ( 74.26,305.66) rectangle ( 80.56,309.32);

\path[fill=fillColor,fill opacity=0.60] ( 80.56,305.66) rectangle ( 86.86,308.14);

\path[fill=fillColor,fill opacity=0.60] ( 86.86,305.66) rectangle ( 93.16,307.30);

\path[fill=fillColor,fill opacity=0.60] ( 93.16,305.66) rectangle ( 99.46,306.94);

\path[fill=fillColor,fill opacity=0.60] ( 99.46,305.66) rectangle (105.76,306.60);

\path[fill=fillColor,fill opacity=0.60] (105.76,305.66) rectangle (112.06,306.36);

\path[fill=fillColor,fill opacity=0.60] (112.06,305.66) rectangle (118.36,306.21);

\path[fill=fillColor,fill opacity=0.60] (118.36,305.66) rectangle (124.66,306.10);

\path[fill=fillColor,fill opacity=0.60] (124.66,305.66) rectangle (130.96,305.99);

\path[fill=fillColor,fill opacity=0.60] (130.96,305.66) rectangle (137.26,305.93);

\path[fill=fillColor,fill opacity=0.60] (137.26,305.66) rectangle (143.56,305.89);

\path[fill=fillColor,fill opacity=0.60] (143.56,305.66) rectangle (149.86,305.90);

\path[fill=fillColor,fill opacity=0.60] (149.86,305.66) rectangle (156.16,305.81);

\path[fill=fillColor,fill opacity=0.60] (156.16,305.66) rectangle (162.46,305.80);

\path[fill=fillColor,fill opacity=0.60] (162.46,305.66) rectangle (168.76,305.78);

\path[fill=fillColor,fill opacity=0.60] (168.76,305.66) rectangle (175.06,305.73);

\path[fill=fillColor,fill opacity=0.60] (175.06,305.66) rectangle (181.36,305.71);

\path[fill=fillColor,fill opacity=0.60] (181.36,305.66) rectangle (187.66,305.72);

\path[fill=fillColor,fill opacity=0.60] (187.66,305.66) rectangle (193.95,305.69);

\path[fill=fillColor,fill opacity=0.60] (193.95,305.66) rectangle (200.25,305.70);

\path[fill=fillColor,fill opacity=0.60] (200.25,305.66) rectangle (206.55,305.69);

\path[fill=fillColor,fill opacity=0.60] (206.55,305.66) rectangle (212.85,305.68);

\path[fill=fillColor,fill opacity=0.60] (212.85,305.66) rectangle (219.15,305.68);
\definecolor{drawColor}{RGB}{0,0,0}

\path[draw=drawColor,line width= 0.6pt,dash pattern=on 4pt off 4pt ,line join=round] ( 33.39,300.36) -- ( 33.39,416.97);

\path[draw=drawColor,line width= 0.6pt,line join=round] ( 34.68,300.36) -- ( 34.68,416.97);

\path[draw=drawColor,line width= 3.4pt,line join=round] ( 33.96,305.66) --
	( 36.29,305.66);

\node[text=drawColor,anchor=base east,inner sep=0pt, outer sep=0pt, scale=  1.00] at (228.60,410.11) {100\%};
\end{scope}
\begin{scope}
\path[clip] (236.85,300.36) rectangle (444.74,416.97);
\definecolor{fillColor}{RGB}{255,255,255}

\path[fill=fillColor] (236.85,300.36) rectangle (444.74,416.97);
\definecolor{fillColor}{RGB}{86,180,233}

\path[fill=fillColor,fill opacity=0.60] (246.30,305.66) rectangle (252.60,306.23);

\path[fill=fillColor,fill opacity=0.60] (252.60,305.66) rectangle (258.90,380.86);

\path[fill=fillColor,fill opacity=0.60] (258.90,305.66) rectangle (265.20,411.67);

\path[fill=fillColor,fill opacity=0.60] (265.20,305.66) rectangle (271.50,375.35);

\path[fill=fillColor,fill opacity=0.60] (271.50,305.66) rectangle (277.80,348.30);

\path[fill=fillColor,fill opacity=0.60] (277.80,305.66) rectangle (284.10,331.73);

\path[fill=fillColor,fill opacity=0.60] (284.10,305.66) rectangle (290.40,322.31);

\path[fill=fillColor,fill opacity=0.60] (290.40,305.66) rectangle (296.70,316.29);

\path[fill=fillColor,fill opacity=0.60] (296.70,305.66) rectangle (303.00,312.65);

\path[fill=fillColor,fill opacity=0.60] (303.00,305.66) rectangle (309.30,310.45);

\path[fill=fillColor,fill opacity=0.60] (309.30,305.66) rectangle (315.60,309.08);

\path[fill=fillColor,fill opacity=0.60] (315.60,305.66) rectangle (321.90,308.11);

\path[fill=fillColor,fill opacity=0.60] (321.90,305.66) rectangle (328.20,307.36);

\path[fill=fillColor,fill opacity=0.60] (328.20,305.66) rectangle (334.50,306.92);

\path[fill=fillColor,fill opacity=0.60] (334.50,305.66) rectangle (340.80,306.67);

\path[fill=fillColor,fill opacity=0.60] (340.80,305.66) rectangle (347.10,306.59);

\path[fill=fillColor,fill opacity=0.60] (347.10,305.66) rectangle (353.40,306.21);

\path[fill=fillColor,fill opacity=0.60] (353.40,305.66) rectangle (359.70,306.15);

\path[fill=fillColor,fill opacity=0.60] (359.70,305.66) rectangle (366.00,306.02);

\path[fill=fillColor,fill opacity=0.60] (366.00,305.66) rectangle (372.30,306.00);

\path[fill=fillColor,fill opacity=0.60] (372.30,305.66) rectangle (378.60,305.94);

\path[fill=fillColor,fill opacity=0.60] (378.60,305.66) rectangle (384.89,305.88);

\path[fill=fillColor,fill opacity=0.60] (384.89,305.66) rectangle (391.19,305.83);

\path[fill=fillColor,fill opacity=0.60] (391.19,305.66) rectangle (397.49,305.83);

\path[fill=fillColor,fill opacity=0.60] (397.49,305.66) rectangle (403.79,305.74);

\path[fill=fillColor,fill opacity=0.60] (403.79,305.66) rectangle (410.09,305.76);

\path[fill=fillColor,fill opacity=0.60] (410.09,305.66) rectangle (416.39,305.72);

\path[fill=fillColor,fill opacity=0.60] (416.39,305.66) rectangle (422.69,305.75);

\path[fill=fillColor,fill opacity=0.60] (422.69,305.66) rectangle (428.99,305.73);

\path[fill=fillColor,fill opacity=0.60] (428.99,305.66) rectangle (435.29,305.67);
\definecolor{drawColor}{RGB}{0,0,0}

\path[draw=drawColor,line width= 0.6pt,dash pattern=on 4pt off 4pt ,line join=round] (249.51,300.36) -- (249.51,416.97);

\path[draw=drawColor,line width= 0.6pt,line join=round] (251.15,300.36) -- (251.15,416.97);

\path[draw=drawColor,line width= 3.4pt,line join=round] (250.30,305.66) --
	(253.33,305.66);

\node[text=drawColor,anchor=base east,inner sep=0pt, outer sep=0pt, scale=  1.00] at (444.74,410.11) {100\%};
\end{scope}
\begin{scope}
\path[clip] (452.99,300.36) rectangle (660.88,416.97);
\definecolor{fillColor}{RGB}{255,255,255}

\path[fill=fillColor] (452.99,300.36) rectangle (660.88,416.97);
\definecolor{fillColor}{RGB}{86,180,233}

\path[fill=fillColor,fill opacity=0.60] (462.44,305.66) rectangle (468.74,305.71);

\path[fill=fillColor,fill opacity=0.60] (468.74,305.66) rectangle (475.04,359.76);

\path[fill=fillColor,fill opacity=0.60] (475.04,305.66) rectangle (481.34,411.67);

\path[fill=fillColor,fill opacity=0.60] (481.34,305.66) rectangle (487.64,380.17);

\path[fill=fillColor,fill opacity=0.60] (487.64,305.66) rectangle (493.94,351.80);

\path[fill=fillColor,fill opacity=0.60] (493.94,305.66) rectangle (500.24,333.70);

\path[fill=fillColor,fill opacity=0.60] (500.24,305.66) rectangle (506.54,322.88);

\path[fill=fillColor,fill opacity=0.60] (506.54,305.66) rectangle (512.84,316.83);

\path[fill=fillColor,fill opacity=0.60] (512.84,305.66) rectangle (519.14,313.04);

\path[fill=fillColor,fill opacity=0.60] (519.14,305.66) rectangle (525.44,310.87);

\path[fill=fillColor,fill opacity=0.60] (525.44,305.66) rectangle (531.74,309.42);

\path[fill=fillColor,fill opacity=0.60] (531.74,305.66) rectangle (538.04,308.32);

\path[fill=fillColor,fill opacity=0.60] (538.04,305.66) rectangle (544.34,307.56);

\path[fill=fillColor,fill opacity=0.60] (544.34,305.66) rectangle (550.64,307.01);

\path[fill=fillColor,fill opacity=0.60] (550.64,305.66) rectangle (556.94,306.77);

\path[fill=fillColor,fill opacity=0.60] (556.94,305.66) rectangle (563.24,306.34);

\path[fill=fillColor,fill opacity=0.60] (563.24,305.66) rectangle (569.54,306.08);

\path[fill=fillColor,fill opacity=0.60] (569.54,305.66) rectangle (575.83,306.25);

\path[fill=fillColor,fill opacity=0.60] (575.83,305.66) rectangle (582.13,306.02);

\path[fill=fillColor,fill opacity=0.60] (582.13,305.66) rectangle (588.43,306.02);

\path[fill=fillColor,fill opacity=0.60] (588.43,305.66) rectangle (594.73,305.86);

\path[fill=fillColor,fill opacity=0.60] (594.73,305.66) rectangle (601.03,305.85);

\path[fill=fillColor,fill opacity=0.60] (601.03,305.66) rectangle (607.33,305.78);

\path[fill=fillColor,fill opacity=0.60] (607.33,305.66) rectangle (613.63,305.81);

\path[fill=fillColor,fill opacity=0.60] (613.63,305.66) rectangle (619.93,305.75);

\path[fill=fillColor,fill opacity=0.60] (619.93,305.66) rectangle (626.23,305.71);

\path[fill=fillColor,fill opacity=0.60] (626.23,305.66) rectangle (632.53,305.71);

\path[fill=fillColor,fill opacity=0.60] (632.53,305.66) rectangle (638.83,305.71);

\path[fill=fillColor,fill opacity=0.60] (638.83,305.66) rectangle (645.13,305.71);

\path[fill=fillColor,fill opacity=0.60] (645.13,305.66) rectangle (651.43,305.67);
\definecolor{drawColor}{RGB}{0,0,0}

\path[draw=drawColor,line width= 0.6pt,dash pattern=on 4pt off 4pt ,line join=round] (465.63,300.36) -- (465.63,416.97);

\path[draw=drawColor,line width= 0.6pt,line join=round] (467.03,300.36) -- (467.03,416.97);

\path[draw=drawColor,line width= 3.4pt,line join=round] (466.44,305.66) --
	(470.12,305.66);

\node[text=drawColor,anchor=base east,inner sep=0pt, outer sep=0pt, scale=  1.00] at (660.88,410.11) {100\%};
\end{scope}
\begin{scope}
\path[clip] ( 20.71,165.52) rectangle (228.60,282.13);
\definecolor{fillColor}{RGB}{255,255,255}

\path[fill=fillColor] ( 20.71,165.52) rectangle (228.60,282.13);
\definecolor{fillColor}{RGB}{204,121,167}

\path[fill=fillColor,fill opacity=0.60] ( 30.16,170.82) rectangle ( 36.46,170.83);

\path[fill=fillColor,fill opacity=0.60] ( 36.46,170.82) rectangle ( 42.76,170.91);

\path[fill=fillColor,fill opacity=0.60] ( 42.76,170.82) rectangle ( 49.06,174.36);

\path[fill=fillColor,fill opacity=0.60] ( 49.06,170.82) rectangle ( 55.36,209.73);

\path[fill=fillColor,fill opacity=0.60] ( 55.36,170.82) rectangle ( 61.66,276.83);

\path[fill=fillColor,fill opacity=0.60] ( 61.66,170.82) rectangle ( 67.96,256.57);

\path[fill=fillColor,fill opacity=0.60] ( 67.96,170.82) rectangle ( 74.26,195.49);

\path[fill=fillColor,fill opacity=0.60] ( 74.26,170.82) rectangle ( 80.56,173.79);

\path[fill=fillColor,fill opacity=0.60] ( 80.56,170.82) rectangle ( 86.86,170.82);

\path[fill=fillColor,fill opacity=0.60] ( 86.86,170.82) rectangle ( 93.16,170.82);

\path[fill=fillColor,fill opacity=0.60] ( 93.16,170.82) rectangle ( 99.46,170.82);

\path[fill=fillColor,fill opacity=0.60] ( 99.46,170.82) rectangle (105.76,170.82);

\path[fill=fillColor,fill opacity=0.60] (105.76,170.82) rectangle (112.06,170.82);

\path[fill=fillColor,fill opacity=0.60] (112.06,170.82) rectangle (118.36,170.82);

\path[fill=fillColor,fill opacity=0.60] (118.36,170.82) rectangle (124.66,170.82);

\path[fill=fillColor,fill opacity=0.60] (124.66,170.82) rectangle (130.96,170.82);

\path[fill=fillColor,fill opacity=0.60] (130.96,170.82) rectangle (137.26,170.82);

\path[fill=fillColor,fill opacity=0.60] (137.26,170.82) rectangle (143.56,170.82);

\path[fill=fillColor,fill opacity=0.60] (143.56,170.82) rectangle (149.86,170.82);

\path[fill=fillColor,fill opacity=0.60] (149.86,170.82) rectangle (156.16,170.82);

\path[fill=fillColor,fill opacity=0.60] (156.16,170.82) rectangle (162.46,170.82);

\path[fill=fillColor,fill opacity=0.60] (162.46,170.82) rectangle (168.76,170.82);

\path[fill=fillColor,fill opacity=0.60] (168.76,170.82) rectangle (175.06,170.82);

\path[fill=fillColor,fill opacity=0.60] (175.06,170.82) rectangle (181.36,170.82);

\path[fill=fillColor,fill opacity=0.60] (181.36,170.82) rectangle (187.66,170.82);

\path[fill=fillColor,fill opacity=0.60] (187.66,170.82) rectangle (193.95,170.82);

\path[fill=fillColor,fill opacity=0.60] (193.95,170.82) rectangle (200.25,170.82);

\path[fill=fillColor,fill opacity=0.60] (200.25,170.82) rectangle (206.55,170.82);

\path[fill=fillColor,fill opacity=0.60] (206.55,170.82) rectangle (212.85,170.82);

\path[fill=fillColor,fill opacity=0.60] (212.85,170.82) rectangle (219.15,170.82);
\definecolor{drawColor}{RGB}{0,0,0}

\path[draw=drawColor,line width= 0.6pt,dash pattern=on 4pt off 4pt ,line join=round] ( 43.39,165.52) -- ( 43.39,282.13);

\path[draw=drawColor,line width= 0.6pt,line join=round] ( 72.58,165.52) -- ( 72.58,282.13);

\path[draw=drawColor,line width= 3.4pt,line join=round] ( 39.18,170.82) --
	(217.93,170.82);

\node[text=drawColor,anchor=base east,inner sep=0pt, outer sep=0pt, scale=  1.00] at (228.60,275.28) {99\%};
\end{scope}
\begin{scope}
\path[clip] (236.85,165.52) rectangle (444.74,282.13);
\definecolor{fillColor}{RGB}{255,255,255}

\path[fill=fillColor] (236.85,165.52) rectangle (444.74,282.13);
\definecolor{fillColor}{RGB}{204,121,167}

\path[fill=fillColor,fill opacity=0.60] (246.30,170.82) rectangle (252.60,170.82);

\path[fill=fillColor,fill opacity=0.60] (252.60,170.82) rectangle (258.90,170.82);

\path[fill=fillColor,fill opacity=0.60] (258.90,170.82) rectangle (265.20,170.82);

\path[fill=fillColor,fill opacity=0.60] (265.20,170.82) rectangle (271.50,170.82);

\path[fill=fillColor,fill opacity=0.60] (271.50,170.82) rectangle (277.80,170.82);

\path[fill=fillColor,fill opacity=0.60] (277.80,170.82) rectangle (284.10,170.82);

\path[fill=fillColor,fill opacity=0.60] (284.10,170.82) rectangle (290.40,170.82);

\path[fill=fillColor,fill opacity=0.60] (290.40,170.82) rectangle (296.70,170.82);

\path[fill=fillColor,fill opacity=0.60] (296.70,170.82) rectangle (303.00,170.83);

\path[fill=fillColor,fill opacity=0.60] (303.00,170.82) rectangle (309.30,170.93);

\path[fill=fillColor,fill opacity=0.60] (309.30,170.82) rectangle (315.60,171.31);

\path[fill=fillColor,fill opacity=0.60] (315.60,170.82) rectangle (321.90,173.04);

\path[fill=fillColor,fill opacity=0.60] (321.90,170.82) rectangle (328.20,177.24);

\path[fill=fillColor,fill opacity=0.60] (328.20,170.82) rectangle (334.50,186.92);

\path[fill=fillColor,fill opacity=0.60] (334.50,170.82) rectangle (340.80,203.82);

\path[fill=fillColor,fill opacity=0.60] (340.80,170.82) rectangle (347.10,226.91);

\path[fill=fillColor,fill opacity=0.60] (347.10,170.82) rectangle (353.40,249.30);

\path[fill=fillColor,fill opacity=0.60] (353.40,170.82) rectangle (359.70,268.73);

\path[fill=fillColor,fill opacity=0.60] (359.70,170.82) rectangle (366.00,276.83);

\path[fill=fillColor,fill opacity=0.60] (366.00,170.82) rectangle (372.30,272.39);

\path[fill=fillColor,fill opacity=0.60] (372.30,170.82) rectangle (378.60,260.21);

\path[fill=fillColor,fill opacity=0.60] (378.60,170.82) rectangle (384.89,241.13);

\path[fill=fillColor,fill opacity=0.60] (384.89,170.82) rectangle (391.19,219.64);

\path[fill=fillColor,fill opacity=0.60] (391.19,170.82) rectangle (397.49,202.21);

\path[fill=fillColor,fill opacity=0.60] (397.49,170.82) rectangle (403.79,190.31);

\path[fill=fillColor,fill opacity=0.60] (403.79,170.82) rectangle (410.09,182.42);

\path[fill=fillColor,fill opacity=0.60] (410.09,170.82) rectangle (416.39,177.15);

\path[fill=fillColor,fill opacity=0.60] (416.39,170.82) rectangle (422.69,174.39);

\path[fill=fillColor,fill opacity=0.60] (422.69,170.82) rectangle (428.99,172.49);

\path[fill=fillColor,fill opacity=0.60] (428.99,170.82) rectangle (435.29,171.66);
\definecolor{drawColor}{RGB}{0,0,0}

\path[draw=drawColor,line width= 0.6pt,dash pattern=on 4pt off 4pt ,line join=round] (267.84,165.52) -- (267.84,282.13);

\path[draw=drawColor,line width= 0.6pt,line join=round] (265.73,165.52) -- (265.73,282.13);

\path[draw=drawColor,line width= 3.4pt,line join=round] (252.05,170.82) --
	(307.57,170.82);

\node[text=drawColor,anchor=base east,inner sep=0pt, outer sep=0pt, scale=  1.00] at (444.74,275.28) {100\%};
\end{scope}
\begin{scope}
\path[clip] (452.99,165.52) rectangle (660.88,282.13);
\definecolor{fillColor}{RGB}{255,255,255}

\path[fill=fillColor] (452.99,165.52) rectangle (660.88,282.13);
\definecolor{fillColor}{RGB}{204,121,167}

\path[fill=fillColor,fill opacity=0.60] (462.44,170.82) rectangle (468.74,170.82);

\path[fill=fillColor,fill opacity=0.60] (468.74,170.82) rectangle (475.04,170.82);

\path[fill=fillColor,fill opacity=0.60] (475.04,170.82) rectangle (481.34,170.82);

\path[fill=fillColor,fill opacity=0.60] (481.34,170.82) rectangle (487.64,170.82);

\path[fill=fillColor,fill opacity=0.60] (487.64,170.82) rectangle (493.94,170.86);

\path[fill=fillColor,fill opacity=0.60] (493.94,170.82) rectangle (500.24,171.06);

\path[fill=fillColor,fill opacity=0.60] (500.24,170.82) rectangle (506.54,172.89);

\path[fill=fillColor,fill opacity=0.60] (506.54,170.82) rectangle (512.84,178.11);

\path[fill=fillColor,fill opacity=0.60] (512.84,170.82) rectangle (519.14,191.74);

\path[fill=fillColor,fill opacity=0.60] (519.14,170.82) rectangle (525.44,213.70);

\path[fill=fillColor,fill opacity=0.60] (525.44,170.82) rectangle (531.74,241.64);

\path[fill=fillColor,fill opacity=0.60] (531.74,170.82) rectangle (538.04,263.48);

\path[fill=fillColor,fill opacity=0.60] (538.04,170.82) rectangle (544.34,276.83);

\path[fill=fillColor,fill opacity=0.60] (544.34,170.82) rectangle (550.64,276.68);

\path[fill=fillColor,fill opacity=0.60] (550.64,170.82) rectangle (556.94,270.41);

\path[fill=fillColor,fill opacity=0.60] (556.94,170.82) rectangle (563.24,255.29);

\path[fill=fillColor,fill opacity=0.60] (563.24,170.82) rectangle (569.54,237.77);

\path[fill=fillColor,fill opacity=0.60] (569.54,170.82) rectangle (575.83,224.22);

\path[fill=fillColor,fill opacity=0.60] (575.83,170.82) rectangle (582.13,207.12);

\path[fill=fillColor,fill opacity=0.60] (582.13,170.82) rectangle (588.43,196.26);

\path[fill=fillColor,fill opacity=0.60] (588.43,170.82) rectangle (594.73,190.12);

\path[fill=fillColor,fill opacity=0.60] (594.73,170.82) rectangle (601.03,183.12);

\path[fill=fillColor,fill opacity=0.60] (601.03,170.82) rectangle (607.33,179.12);

\path[fill=fillColor,fill opacity=0.60] (607.33,170.82) rectangle (613.63,175.13);

\path[fill=fillColor,fill opacity=0.60] (613.63,170.82) rectangle (619.93,174.42);

\path[fill=fillColor,fill opacity=0.60] (619.93,170.82) rectangle (626.23,172.95);

\path[fill=fillColor,fill opacity=0.60] (626.23,170.82) rectangle (632.53,172.26);

\path[fill=fillColor,fill opacity=0.60] (632.53,170.82) rectangle (638.83,171.67);

\path[fill=fillColor,fill opacity=0.60] (638.83,170.82) rectangle (645.13,171.41);

\path[fill=fillColor,fill opacity=0.60] (645.13,170.82) rectangle (651.43,171.18);
\definecolor{drawColor}{RGB}{0,0,0}

\path[draw=drawColor,line width= 0.6pt,dash pattern=on 4pt off 4pt ,line join=round] (466.63,165.52) -- (466.63,282.13);

\path[draw=drawColor,line width= 0.6pt,line join=round] (475.29,165.52) -- (475.29,282.13);

\path[draw=drawColor,line width= 3.4pt,line join=round] (467.72,170.82) --
	(508.41,170.82);

\node[text=drawColor,anchor=base east,inner sep=0pt, outer sep=0pt, scale=  1.00] at (660.88,275.28) {100\%};
\end{scope}
\begin{scope}
\path[clip] ( 20.71, 30.69) rectangle (228.60,147.30);
\definecolor{fillColor}{RGB}{255,255,255}

\path[fill=fillColor] ( 20.71, 30.69) rectangle (228.60,147.30);
\definecolor{fillColor}{RGB}{230,159,0}

\path[fill=fillColor,fill opacity=0.60] ( 30.16, 35.99) rectangle ( 36.46, 36.05);

\path[fill=fillColor,fill opacity=0.60] ( 36.46, 35.99) rectangle ( 42.76, 36.09);

\path[fill=fillColor,fill opacity=0.60] ( 42.76, 35.99) rectangle ( 49.06, 36.14);

\path[fill=fillColor,fill opacity=0.60] ( 49.06, 35.99) rectangle ( 55.36, 36.39);

\path[fill=fillColor,fill opacity=0.60] ( 55.36, 35.99) rectangle ( 61.66, 36.94);

\path[fill=fillColor,fill opacity=0.60] ( 61.66, 35.99) rectangle ( 67.96, 38.13);

\path[fill=fillColor,fill opacity=0.60] ( 67.96, 35.99) rectangle ( 74.26, 39.90);

\path[fill=fillColor,fill opacity=0.60] ( 74.26, 35.99) rectangle ( 80.56, 43.67);

\path[fill=fillColor,fill opacity=0.60] ( 80.56, 35.99) rectangle ( 86.86, 48.84);

\path[fill=fillColor,fill opacity=0.60] ( 86.86, 35.99) rectangle ( 93.16, 56.86);

\path[fill=fillColor,fill opacity=0.60] ( 93.16, 35.99) rectangle ( 99.46, 67.28);

\path[fill=fillColor,fill opacity=0.60] ( 99.46, 35.99) rectangle (105.76, 79.37);

\path[fill=fillColor,fill opacity=0.60] (105.76, 35.99) rectangle (112.06, 95.07);

\path[fill=fillColor,fill opacity=0.60] (112.06, 35.99) rectangle (118.36,109.74);

\path[fill=fillColor,fill opacity=0.60] (118.36, 35.99) rectangle (124.66,127.62);

\path[fill=fillColor,fill opacity=0.60] (124.66, 35.99) rectangle (130.96,139.36);

\path[fill=fillColor,fill opacity=0.60] (130.96, 35.99) rectangle (137.26,142.00);

\path[fill=fillColor,fill opacity=0.60] (137.26, 35.99) rectangle (143.56,139.23);

\path[fill=fillColor,fill opacity=0.60] (143.56, 35.99) rectangle (149.86,134.05);

\path[fill=fillColor,fill opacity=0.60] (149.86, 35.99) rectangle (156.16,119.97);

\path[fill=fillColor,fill opacity=0.60] (156.16, 35.99) rectangle (162.46,101.66);

\path[fill=fillColor,fill opacity=0.60] (162.46, 35.99) rectangle (168.76, 85.47);

\path[fill=fillColor,fill opacity=0.60] (168.76, 35.99) rectangle (175.06, 71.31);

\path[fill=fillColor,fill opacity=0.60] (175.06, 35.99) rectangle (181.36, 59.05);

\path[fill=fillColor,fill opacity=0.60] (181.36, 35.99) rectangle (187.66, 49.21);

\path[fill=fillColor,fill opacity=0.60] (187.66, 35.99) rectangle (193.95, 43.92);

\path[fill=fillColor,fill opacity=0.60] (193.95, 35.99) rectangle (200.25, 40.98);

\path[fill=fillColor,fill opacity=0.60] (200.25, 35.99) rectangle (206.55, 38.53);

\path[fill=fillColor,fill opacity=0.60] (206.55, 35.99) rectangle (212.85, 37.28);

\path[fill=fillColor,fill opacity=0.60] (212.85, 35.99) rectangle (219.15, 36.08);
\definecolor{drawColor}{RGB}{0,0,0}

\path[draw=drawColor,line width= 0.6pt,dash pattern=on 4pt off 4pt ,line join=round] (131.62, 30.69) -- (131.62,147.30);

\node[text=drawColor,anchor=base east,inner sep=0pt, outer sep=0pt, scale=  1.00] at (228.60,140.44) {55\%};
\end{scope}
\begin{scope}
\path[clip] (236.85, 30.69) rectangle (444.74,147.30);
\definecolor{fillColor}{RGB}{255,255,255}

\path[fill=fillColor] (236.85, 30.69) rectangle (444.74,147.30);
\definecolor{fillColor}{RGB}{230,159,0}

\path[fill=fillColor,fill opacity=0.60] (246.30, 35.99) rectangle (252.60, 36.00);

\path[fill=fillColor,fill opacity=0.60] (252.60, 35.99) rectangle (258.90, 36.03);

\path[fill=fillColor,fill opacity=0.60] (258.90, 35.99) rectangle (265.20, 36.18);

\path[fill=fillColor,fill opacity=0.60] (265.20, 35.99) rectangle (271.50, 36.45);

\path[fill=fillColor,fill opacity=0.60] (271.50, 35.99) rectangle (277.80, 37.45);

\path[fill=fillColor,fill opacity=0.60] (277.80, 35.99) rectangle (284.10, 39.53);

\path[fill=fillColor,fill opacity=0.60] (284.10, 35.99) rectangle (290.40, 44.38);

\path[fill=fillColor,fill opacity=0.60] (290.40, 35.99) rectangle (296.70, 53.11);

\path[fill=fillColor,fill opacity=0.60] (296.70, 35.99) rectangle (303.00, 65.90);

\path[fill=fillColor,fill opacity=0.60] (303.00, 35.99) rectangle (309.30, 82.76);

\path[fill=fillColor,fill opacity=0.60] (309.30, 35.99) rectangle (315.60,105.93);

\path[fill=fillColor,fill opacity=0.60] (315.60, 35.99) rectangle (321.90,126.96);

\path[fill=fillColor,fill opacity=0.60] (321.90, 35.99) rectangle (328.20,138.50);

\path[fill=fillColor,fill opacity=0.60] (328.20, 35.99) rectangle (334.50,142.00);

\path[fill=fillColor,fill opacity=0.60] (334.50, 35.99) rectangle (340.80,137.38);

\path[fill=fillColor,fill opacity=0.60] (340.80, 35.99) rectangle (347.10,125.99);

\path[fill=fillColor,fill opacity=0.60] (347.10, 35.99) rectangle (353.40,112.21);

\path[fill=fillColor,fill opacity=0.60] (353.40, 35.99) rectangle (359.70,100.37);

\path[fill=fillColor,fill opacity=0.60] (359.70, 35.99) rectangle (366.00, 90.24);

\path[fill=fillColor,fill opacity=0.60] (366.00, 35.99) rectangle (372.30, 79.86);

\path[fill=fillColor,fill opacity=0.60] (372.30, 35.99) rectangle (378.60, 71.83);

\path[fill=fillColor,fill opacity=0.60] (378.60, 35.99) rectangle (384.89, 64.10);

\path[fill=fillColor,fill opacity=0.60] (384.89, 35.99) rectangle (391.19, 58.43);

\path[fill=fillColor,fill opacity=0.60] (391.19, 35.99) rectangle (397.49, 51.57);

\path[fill=fillColor,fill opacity=0.60] (397.49, 35.99) rectangle (403.79, 46.78);

\path[fill=fillColor,fill opacity=0.60] (403.79, 35.99) rectangle (410.09, 42.94);

\path[fill=fillColor,fill opacity=0.60] (410.09, 35.99) rectangle (416.39, 40.83);

\path[fill=fillColor,fill opacity=0.60] (416.39, 35.99) rectangle (422.69, 38.76);

\path[fill=fillColor,fill opacity=0.60] (422.69, 35.99) rectangle (428.99, 37.40);

\path[fill=fillColor,fill opacity=0.60] (428.99, 35.99) rectangle (435.29, 36.75);
\definecolor{drawColor}{RGB}{0,0,0}

\path[draw=drawColor,line width= 0.6pt,dash pattern=on 4pt off 4pt ,line join=round] (310.83, 30.69) -- (310.83,147.30);

\node[text=drawColor,anchor=base east,inner sep=0pt, outer sep=0pt, scale=  1.00] at (444.74,140.44) {88\%};
\end{scope}
\begin{scope}
\path[clip] (452.99, 30.69) rectangle (660.88,147.30);
\definecolor{fillColor}{RGB}{255,255,255}

\path[fill=fillColor] (452.99, 30.69) rectangle (660.88,147.30);
\definecolor{fillColor}{RGB}{230,159,0}

\path[fill=fillColor,fill opacity=0.60] (462.44, 35.99) rectangle (468.74, 36.09);

\path[fill=fillColor,fill opacity=0.60] (468.74, 35.99) rectangle (475.04, 36.22);

\path[fill=fillColor,fill opacity=0.60] (475.04, 35.99) rectangle (481.34, 36.55);

\path[fill=fillColor,fill opacity=0.60] (481.34, 35.99) rectangle (487.64, 36.76);

\path[fill=fillColor,fill opacity=0.60] (487.64, 35.99) rectangle (493.94, 37.99);

\path[fill=fillColor,fill opacity=0.60] (493.94, 35.99) rectangle (500.24, 41.31);

\path[fill=fillColor,fill opacity=0.60] (500.24, 35.99) rectangle (506.54, 44.04);

\path[fill=fillColor,fill opacity=0.60] (506.54, 35.99) rectangle (512.84, 51.45);

\path[fill=fillColor,fill opacity=0.60] (512.84, 35.99) rectangle (519.14, 60.57);

\path[fill=fillColor,fill opacity=0.60] (519.14, 35.99) rectangle (525.44, 75.96);

\path[fill=fillColor,fill opacity=0.60] (525.44, 35.99) rectangle (531.74, 89.21);

\path[fill=fillColor,fill opacity=0.60] (531.74, 35.99) rectangle (538.04,104.58);

\path[fill=fillColor,fill opacity=0.60] (538.04, 35.99) rectangle (544.34,121.90);

\path[fill=fillColor,fill opacity=0.60] (544.34, 35.99) rectangle (550.64,128.80);

\path[fill=fillColor,fill opacity=0.60] (550.64, 35.99) rectangle (556.94,140.71);

\path[fill=fillColor,fill opacity=0.60] (556.94, 35.99) rectangle (563.24,142.00);

\path[fill=fillColor,fill opacity=0.60] (563.24, 35.99) rectangle (569.54,139.24);

\path[fill=fillColor,fill opacity=0.60] (569.54, 35.99) rectangle (575.83,133.45);

\path[fill=fillColor,fill opacity=0.60] (575.83, 35.99) rectangle (582.13,119.22);

\path[fill=fillColor,fill opacity=0.60] (582.13, 35.99) rectangle (588.43,105.35);

\path[fill=fillColor,fill opacity=0.60] (588.43, 35.99) rectangle (594.73, 88.98);

\path[fill=fillColor,fill opacity=0.60] (594.73, 35.99) rectangle (601.03, 75.62);

\path[fill=fillColor,fill opacity=0.60] (601.03, 35.99) rectangle (607.33, 64.94);

\path[fill=fillColor,fill opacity=0.60] (607.33, 35.99) rectangle (613.63, 53.95);

\path[fill=fillColor,fill opacity=0.60] (613.63, 35.99) rectangle (619.93, 48.44);

\path[fill=fillColor,fill opacity=0.60] (619.93, 35.99) rectangle (626.23, 44.43);

\path[fill=fillColor,fill opacity=0.60] (626.23, 35.99) rectangle (632.53, 40.08);

\path[fill=fillColor,fill opacity=0.60] (632.53, 35.99) rectangle (638.83, 38.97);

\path[fill=fillColor,fill opacity=0.60] (638.83, 35.99) rectangle (645.13, 37.58);

\path[fill=fillColor,fill opacity=0.60] (645.13, 35.99) rectangle (651.43, 36.50);
\definecolor{drawColor}{RGB}{0,0,0}

\path[draw=drawColor,line width= 0.6pt,dash pattern=on 4pt off 4pt ,line join=round] (524.04, 30.69) -- (524.04,147.30);

\node[text=drawColor,anchor=base east,inner sep=0pt, outer sep=0pt, scale=  1.00] at (660.88,140.44) {92\%};
\end{scope}
\begin{scope}
\path[clip] ( 20.71,416.97) rectangle (228.60,433.54);
\definecolor{drawColor}{RGB}{0,0,0}
\definecolor{fillColor}{RGB}{255,255,255}

\path[draw=drawColor,line width= 1.1pt,line join=round,line cap=round,fill=fillColor] ( 20.71,416.97) rectangle (228.60,433.54);
\definecolor{drawColor}{gray}{0.10}

\node[text=drawColor,anchor=base,inner sep=0pt, outer sep=0pt, scale=  0.88] at (124.66,422.22) {Austronesian};
\end{scope}
\begin{scope}
\path[clip] (236.85,416.97) rectangle (444.74,433.54);
\definecolor{drawColor}{RGB}{0,0,0}
\definecolor{fillColor}{RGB}{255,255,255}

\path[draw=drawColor,line width= 1.1pt,line join=round,line cap=round,fill=fillColor] (236.85,416.97) rectangle (444.74,433.54);
\definecolor{drawColor}{gray}{0.10}

\node[text=drawColor,anchor=base,inner sep=0pt, outer sep=0pt, scale=  0.88] at (340.80,422.22) {Semitic};
\end{scope}
\begin{scope}
\path[clip] (452.99,416.97) rectangle (660.88,433.54);
\definecolor{drawColor}{RGB}{0,0,0}
\definecolor{fillColor}{RGB}{255,255,255}

\path[draw=drawColor,line width= 1.1pt,line join=round,line cap=round,fill=fillColor] (452.99,416.97) rectangle (660.88,433.54);
\definecolor{drawColor}{gray}{0.10}

\node[text=drawColor,anchor=base,inner sep=0pt, outer sep=0pt, scale=  0.88] at (556.94,422.22) {Uralic};
\end{scope}
\begin{scope}
\path[clip] (660.88,300.36) rectangle (677.45,416.97);
\definecolor{drawColor}{RGB}{0,0,0}
\definecolor{fillColor}{RGB}{255,255,255}

\path[draw=drawColor,line width= 1.1pt,line join=round,line cap=round,fill=fillColor] (660.88,300.36) rectangle (677.45,416.97);
\definecolor{drawColor}{gray}{0.10}

\node[text=drawColor,rotate=-90.00,anchor=base,inner sep=0pt, outer sep=0pt, scale=  0.88] at (666.14,358.66) {Birth rate, $-$IC vs.\ $+$IC};
\end{scope}
\begin{scope}
\path[clip] (660.88,165.52) rectangle (677.45,282.13);
\definecolor{drawColor}{RGB}{0,0,0}
\definecolor{fillColor}{RGB}{255,255,255}

\path[draw=drawColor,line width= 1.1pt,line join=round,line cap=round,fill=fillColor] (660.88,165.52) rectangle (677.45,282.13);
\definecolor{drawColor}{gray}{0.10}

\node[text=drawColor,rotate=-90.00,anchor=base,inner sep=0pt, outer sep=0pt, scale=  0.88] at (666.14,223.83) {Mutation rate to $-$IC vs.\ $+$IC};
\end{scope}
\begin{scope}
\path[clip] (660.88, 30.69) rectangle (677.45,147.30);
\definecolor{drawColor}{RGB}{0,0,0}
\definecolor{fillColor}{RGB}{255,255,255}

\path[draw=drawColor,line width= 1.1pt,line join=round,line cap=round,fill=fillColor] (660.88, 30.69) rectangle (677.45,147.30);
\definecolor{drawColor}{gray}{0.10}

\node[text=drawColor,rotate=-90.00,anchor=base,inner sep=0pt, outer sep=0pt, scale=  0.88] at (666.14, 88.99) {Loss rate, $+$IC vs.\ $-$IC};
\end{scope}
\begin{scope}
\path[clip] (  0.00,  0.00) rectangle (682.95,439.04);
\definecolor{drawColor}{RGB}{0,0,0}

\path[draw=drawColor,line width= 0.6pt,line join=round] ( 20.71, 30.69) --
	(228.60, 30.69);
\end{scope}
\begin{scope}
\path[clip] (  0.00,  0.00) rectangle (682.95,439.04);
\definecolor{drawColor}{gray}{0.20}

\path[draw=drawColor,line width= 0.6pt,line join=round] ( 37.62, 27.94) --
	( 37.62, 30.69);

\path[draw=drawColor,line width= 0.6pt,line join=round] ( 84.62, 27.94) --
	( 84.62, 30.69);

\path[draw=drawColor,line width= 0.6pt,line join=round] (131.62, 27.94) --
	(131.62, 30.69);

\path[draw=drawColor,line width= 0.6pt,line join=round] (178.62, 27.94) --
	(178.62, 30.69);

\path[draw=drawColor,line width= 0.6pt,line join=round] (225.61, 27.94) --
	(225.61, 30.69);
\end{scope}
\begin{scope}
\path[clip] (  0.00,  0.00) rectangle (682.95,439.04);
\definecolor{drawColor}{gray}{0.30}

\node[text=drawColor,anchor=base,inner sep=0pt, outer sep=0pt, scale=  0.88] at ( 37.62, 19.68) {0.8};

\node[text=drawColor,anchor=base,inner sep=0pt, outer sep=0pt, scale=  0.88] at ( 84.62, 19.68) {0.9};

\node[text=drawColor,anchor=base,inner sep=0pt, outer sep=0pt, scale=  0.88] at (131.62, 19.68) {1.0};

\node[text=drawColor,anchor=base,inner sep=0pt, outer sep=0pt, scale=  0.88] at (178.62, 19.68) {1.1};

\node[text=drawColor,anchor=base,inner sep=0pt, outer sep=0pt, scale=  0.88] at (225.61, 19.68) {1.2};
\end{scope}
\begin{scope}
\path[clip] (  0.00,  0.00) rectangle (682.95,439.04);
\definecolor{drawColor}{RGB}{0,0,0}

\path[draw=drawColor,line width= 0.6pt,line join=round] (236.85, 30.69) --
	(444.74, 30.69);
\end{scope}
\begin{scope}
\path[clip] (  0.00,  0.00) rectangle (682.95,439.04);
\definecolor{drawColor}{gray}{0.20}

\path[draw=drawColor,line width= 0.6pt,line join=round] (264.54, 27.94) --
	(264.54, 30.69);

\path[draw=drawColor,line width= 0.6pt,line join=round] (310.83, 27.94) --
	(310.83, 30.69);

\path[draw=drawColor,line width= 0.6pt,line join=round] (357.13, 27.94) --
	(357.13, 30.69);

\path[draw=drawColor,line width= 0.6pt,line join=round] (403.42, 27.94) --
	(403.42, 30.69);
\end{scope}
\begin{scope}
\path[clip] (  0.00,  0.00) rectangle (682.95,439.04);
\definecolor{drawColor}{gray}{0.30}

\node[text=drawColor,anchor=base,inner sep=0pt, outer sep=0pt, scale=  0.88] at (264.54, 19.68) {0.8};

\node[text=drawColor,anchor=base,inner sep=0pt, outer sep=0pt, scale=  0.88] at (310.83, 19.68) {1.0};

\node[text=drawColor,anchor=base,inner sep=0pt, outer sep=0pt, scale=  0.88] at (357.13, 19.68) {1.2};

\node[text=drawColor,anchor=base,inner sep=0pt, outer sep=0pt, scale=  0.88] at (403.42, 19.68) {1.4};
\end{scope}
\begin{scope}
\path[clip] (  0.00,  0.00) rectangle (682.95,439.04);
\definecolor{drawColor}{RGB}{0,0,0}

\path[draw=drawColor,line width= 0.6pt,line join=round] (452.99, 30.69) --
	(660.88, 30.69);
\end{scope}
\begin{scope}
\path[clip] (  0.00,  0.00) rectangle (682.95,439.04);
\definecolor{drawColor}{gray}{0.20}

\path[draw=drawColor,line width= 0.6pt,line join=round] (478.33, 27.94) --
	(478.33, 30.69);

\path[draw=drawColor,line width= 0.6pt,line join=round] (524.04, 27.94) --
	(524.04, 30.69);

\path[draw=drawColor,line width= 0.6pt,line join=round] (569.75, 27.94) --
	(569.75, 30.69);

\path[draw=drawColor,line width= 0.6pt,line join=round] (615.46, 27.94) --
	(615.46, 30.69);
\end{scope}
\begin{scope}
\path[clip] (  0.00,  0.00) rectangle (682.95,439.04);
\definecolor{drawColor}{gray}{0.30}

\node[text=drawColor,anchor=base,inner sep=0pt, outer sep=0pt, scale=  0.88] at (478.33, 19.68) {0.8};

\node[text=drawColor,anchor=base,inner sep=0pt, outer sep=0pt, scale=  0.88] at (524.04, 19.68) {1.0};

\node[text=drawColor,anchor=base,inner sep=0pt, outer sep=0pt, scale=  0.88] at (569.75, 19.68) {1.2};

\node[text=drawColor,anchor=base,inner sep=0pt, outer sep=0pt, scale=  0.88] at (615.46, 19.68) {1.4};
\end{scope}
\begin{scope}
\path[clip] (  0.00,  0.00) rectangle (682.95,439.04);
\definecolor{drawColor}{RGB}{0,0,0}

\path[draw=drawColor,line width= 0.6pt,line join=round] ( 20.71,165.52) --
	(228.60,165.52);
\end{scope}
\begin{scope}
\path[clip] (  0.00,  0.00) rectangle (682.95,439.04);
\definecolor{drawColor}{gray}{0.20}

\path[draw=drawColor,line width= 0.6pt,line join=round] ( 72.36,162.77) --
	( 72.36,165.52);

\path[draw=drawColor,line width= 0.6pt,line join=round] (130.30,162.77) --
	(130.30,165.52);

\path[draw=drawColor,line width= 0.6pt,line join=round] (188.24,162.77) --
	(188.24,165.52);
\end{scope}
\begin{scope}
\path[clip] (  0.00,  0.00) rectangle (682.95,439.04);
\definecolor{drawColor}{gray}{0.30}

\node[text=drawColor,anchor=base,inner sep=0pt, outer sep=0pt, scale=  0.88] at ( 72.36,154.51) {2};

\node[text=drawColor,anchor=base,inner sep=0pt, outer sep=0pt, scale=  0.88] at (130.30,154.51) {4};

\node[text=drawColor,anchor=base,inner sep=0pt, outer sep=0pt, scale=  0.88] at (188.24,154.51) {6};
\end{scope}
\begin{scope}
\path[clip] (  0.00,  0.00) rectangle (682.95,439.04);
\definecolor{drawColor}{RGB}{0,0,0}

\path[draw=drawColor,line width= 0.6pt,line join=round] (236.85,165.52) --
	(444.74,165.52);
\end{scope}
\begin{scope}
\path[clip] (  0.00,  0.00) rectangle (682.95,439.04);
\definecolor{drawColor}{gray}{0.20}

\path[draw=drawColor,line width= 0.6pt,line join=round] (267.84,162.77) --
	(267.84,165.52);

\path[draw=drawColor,line width= 0.6pt,line join=round] (305.38,162.77) --
	(305.38,165.52);

\path[draw=drawColor,line width= 0.6pt,line join=round] (342.92,162.77) --
	(342.92,165.52);

\path[draw=drawColor,line width= 0.6pt,line join=round] (380.46,162.77) --
	(380.46,165.52);

\path[draw=drawColor,line width= 0.6pt,line join=round] (418.00,162.77) --
	(418.00,165.52);
\end{scope}
\begin{scope}
\path[clip] (  0.00,  0.00) rectangle (682.95,439.04);
\definecolor{drawColor}{gray}{0.30}

\node[text=drawColor,anchor=base,inner sep=0pt, outer sep=0pt, scale=  0.88] at (267.84,154.51) {1.0};

\node[text=drawColor,anchor=base,inner sep=0pt, outer sep=0pt, scale=  0.88] at (305.38,154.51) {1.5};

\node[text=drawColor,anchor=base,inner sep=0pt, outer sep=0pt, scale=  0.88] at (342.92,154.51) {2.0};

\node[text=drawColor,anchor=base,inner sep=0pt, outer sep=0pt, scale=  0.88] at (380.46,154.51) {2.5};

\node[text=drawColor,anchor=base,inner sep=0pt, outer sep=0pt, scale=  0.88] at (418.00,154.51) {3.0};
\end{scope}
\begin{scope}
\path[clip] (  0.00,  0.00) rectangle (682.95,439.04);
\definecolor{drawColor}{RGB}{0,0,0}

\path[draw=drawColor,line width= 0.6pt,line join=round] (452.99,165.52) --
	(660.88,165.52);
\end{scope}
\begin{scope}
\path[clip] (  0.00,  0.00) rectangle (682.95,439.04);
\definecolor{drawColor}{gray}{0.20}

\path[draw=drawColor,line width= 0.6pt,line join=round] (452.99,162.77) --
	(452.99,165.52);

\path[draw=drawColor,line width= 0.6pt,line join=round] (521.19,162.77) --
	(521.19,165.52);

\path[draw=drawColor,line width= 0.6pt,line join=round] (589.38,162.77) --
	(589.38,165.52);

\path[draw=drawColor,line width= 0.6pt,line join=round] (657.58,162.77) --
	(657.58,165.52);
\end{scope}
\begin{scope}
\path[clip] (  0.00,  0.00) rectangle (682.95,439.04);
\definecolor{drawColor}{gray}{0.30}

\node[text=drawColor,anchor=base,inner sep=0pt, outer sep=0pt, scale=  0.88] at (452.99,154.51) {0};

\node[text=drawColor,anchor=base,inner sep=0pt, outer sep=0pt, scale=  0.88] at (521.19,154.51) {5};

\node[text=drawColor,anchor=base,inner sep=0pt, outer sep=0pt, scale=  0.88] at (589.38,154.51) {10};

\node[text=drawColor,anchor=base,inner sep=0pt, outer sep=0pt, scale=  0.88] at (657.58,154.51) {15};
\end{scope}
\begin{scope}
\path[clip] (  0.00,  0.00) rectangle (682.95,439.04);
\definecolor{drawColor}{RGB}{0,0,0}

\path[draw=drawColor,line width= 0.6pt,line join=round] ( 20.71,300.36) --
	(228.60,300.36);
\end{scope}
\begin{scope}
\path[clip] (  0.00,  0.00) rectangle (682.95,439.04);
\definecolor{drawColor}{gray}{0.20}

\path[draw=drawColor,line width= 0.6pt,line join=round] ( 33.31,297.61) --
	( 33.31,300.36);

\path[draw=drawColor,line width= 0.6pt,line join=round] ( 73.58,297.61) --
	( 73.58,300.36);

\path[draw=drawColor,line width= 0.6pt,line join=round] (113.85,297.61) --
	(113.85,300.36);

\path[draw=drawColor,line width= 0.6pt,line join=round] (154.12,297.61) --
	(154.12,300.36);

\path[draw=drawColor,line width= 0.6pt,line join=round] (194.39,297.61) --
	(194.39,300.36);
\end{scope}
\begin{scope}
\path[clip] (  0.00,  0.00) rectangle (682.95,439.04);
\definecolor{drawColor}{gray}{0.30}

\node[text=drawColor,anchor=base,inner sep=0pt, outer sep=0pt, scale=  0.88] at ( 33.31,289.34) {0};

\node[text=drawColor,anchor=base,inner sep=0pt, outer sep=0pt, scale=  0.88] at ( 73.58,289.34) {500};

\node[text=drawColor,anchor=base,inner sep=0pt, outer sep=0pt, scale=  0.88] at (113.85,289.34) {1000};

\node[text=drawColor,anchor=base,inner sep=0pt, outer sep=0pt, scale=  0.88] at (154.12,289.34) {1500};

\node[text=drawColor,anchor=base,inner sep=0pt, outer sep=0pt, scale=  0.88] at (194.39,289.34) {2000};
\end{scope}
\begin{scope}
\path[clip] (  0.00,  0.00) rectangle (682.95,439.04);
\definecolor{drawColor}{RGB}{0,0,0}

\path[draw=drawColor,line width= 0.6pt,line join=round] (236.85,300.36) --
	(444.74,300.36);
\end{scope}
\begin{scope}
\path[clip] (  0.00,  0.00) rectangle (682.95,439.04);
\definecolor{drawColor}{gray}{0.20}

\path[draw=drawColor,line width= 0.6pt,line join=round] (249.45,297.61) --
	(249.45,300.36);

\path[draw=drawColor,line width= 0.6pt,line join=round] (302.64,297.61) --
	(302.64,300.36);

\path[draw=drawColor,line width= 0.6pt,line join=round] (355.83,297.61) --
	(355.83,300.36);

\path[draw=drawColor,line width= 0.6pt,line join=round] (409.02,297.61) --
	(409.02,300.36);
\end{scope}
\begin{scope}
\path[clip] (  0.00,  0.00) rectangle (682.95,439.04);
\definecolor{drawColor}{gray}{0.30}

\node[text=drawColor,anchor=base,inner sep=0pt, outer sep=0pt, scale=  0.88] at (249.45,289.34) {0};

\node[text=drawColor,anchor=base,inner sep=0pt, outer sep=0pt, scale=  0.88] at (302.64,289.34) {1000};

\node[text=drawColor,anchor=base,inner sep=0pt, outer sep=0pt, scale=  0.88] at (355.83,289.34) {2000};

\node[text=drawColor,anchor=base,inner sep=0pt, outer sep=0pt, scale=  0.88] at (409.02,289.34) {3000};
\end{scope}
\begin{scope}
\path[clip] (  0.00,  0.00) rectangle (682.95,439.04);
\definecolor{drawColor}{RGB}{0,0,0}

\path[draw=drawColor,line width= 0.6pt,line join=round] (452.99,300.36) --
	(660.88,300.36);
\end{scope}
\begin{scope}
\path[clip] (  0.00,  0.00) rectangle (682.95,439.04);
\definecolor{drawColor}{gray}{0.20}

\path[draw=drawColor,line width= 0.6pt,line join=round] (465.59,297.61) --
	(465.59,300.36);

\path[draw=drawColor,line width= 0.6pt,line join=round] (506.04,297.61) --
	(506.04,300.36);

\path[draw=drawColor,line width= 0.6pt,line join=round] (546.48,297.61) --
	(546.48,300.36);

\path[draw=drawColor,line width= 0.6pt,line join=round] (586.93,297.61) --
	(586.93,300.36);

\path[draw=drawColor,line width= 0.6pt,line join=round] (627.38,297.61) --
	(627.38,300.36);
\end{scope}
\begin{scope}
\path[clip] (  0.00,  0.00) rectangle (682.95,439.04);
\definecolor{drawColor}{gray}{0.30}

\node[text=drawColor,anchor=base,inner sep=0pt, outer sep=0pt, scale=  0.88] at (465.59,289.34) {0};

\node[text=drawColor,anchor=base,inner sep=0pt, outer sep=0pt, scale=  0.88] at (506.04,289.34) {1000};

\node[text=drawColor,anchor=base,inner sep=0pt, outer sep=0pt, scale=  0.88] at (546.48,289.34) {2000};

\node[text=drawColor,anchor=base,inner sep=0pt, outer sep=0pt, scale=  0.88] at (586.93,289.34) {3000};

\node[text=drawColor,anchor=base,inner sep=0pt, outer sep=0pt, scale=  0.88] at (627.38,289.34) {4000};
\end{scope}
\begin{scope}
\path[clip] (  0.00,  0.00) rectangle (682.95,439.04);
\definecolor{drawColor}{RGB}{0,0,0}

\path[draw=drawColor,line width= 0.6pt,line join=round] (452.99,300.36) --
	(452.99,416.97);
\end{scope}
\begin{scope}
\path[clip] (  0.00,  0.00) rectangle (682.95,439.04);
\definecolor{drawColor}{RGB}{0,0,0}

\path[draw=drawColor,line width= 0.6pt,line join=round] (452.99,165.52) --
	(452.99,282.13);
\end{scope}
\begin{scope}
\path[clip] (  0.00,  0.00) rectangle (682.95,439.04);
\definecolor{drawColor}{RGB}{0,0,0}

\path[draw=drawColor,line width= 0.6pt,line join=round] (452.99, 30.69) --
	(452.99,147.30);
\end{scope}
\begin{scope}
\path[clip] (  0.00,  0.00) rectangle (682.95,439.04);
\definecolor{drawColor}{RGB}{0,0,0}

\path[draw=drawColor,line width= 0.6pt,line join=round] (236.85,300.36) --
	(236.85,416.97);
\end{scope}
\begin{scope}
\path[clip] (  0.00,  0.00) rectangle (682.95,439.04);
\definecolor{drawColor}{RGB}{0,0,0}

\path[draw=drawColor,line width= 0.6pt,line join=round] (236.85,165.52) --
	(236.85,282.13);
\end{scope}
\begin{scope}
\path[clip] (  0.00,  0.00) rectangle (682.95,439.04);
\definecolor{drawColor}{RGB}{0,0,0}

\path[draw=drawColor,line width= 0.6pt,line join=round] (236.85, 30.69) --
	(236.85,147.30);
\end{scope}
\begin{scope}
\path[clip] (  0.00,  0.00) rectangle (682.95,439.04);
\definecolor{drawColor}{RGB}{0,0,0}

\path[draw=drawColor,line width= 0.6pt,line join=round] ( 20.71,300.36) --
	( 20.71,416.97);
\end{scope}
\begin{scope}
\path[clip] (  0.00,  0.00) rectangle (682.95,439.04);
\definecolor{drawColor}{RGB}{0,0,0}

\path[draw=drawColor,line width= 0.6pt,line join=round] ( 20.71,165.52) --
	( 20.71,282.13);
\end{scope}
\begin{scope}
\path[clip] (  0.00,  0.00) rectangle (682.95,439.04);
\definecolor{drawColor}{RGB}{0,0,0}

\path[draw=drawColor,line width= 0.6pt,line join=round] ( 20.71, 30.69) --
	( 20.71,147.30);
\end{scope}
\begin{scope}
\path[clip] (  0.00,  0.00) rectangle (682.95,439.04);
\definecolor{drawColor}{RGB}{0,0,0}

\node[text=drawColor,rotate= 90.00,anchor=base,inner sep=0pt, outer sep=0pt, scale=  1.10] at ( 13.08,223.83) {Posterior density};
\end{scope}
\end{tikzpicture}

%% file: discussion.tex
\section{Discussion}
Various aspects of linguistic systems, including 
lexical items, 
are argued to be optimized for communicative efficiency \cite{regier2007,graff2012communicative,ferrer2022optimal}. 
To date, little work has explicitly explored the evolutionary dynamics that give rise to efficient systems, with some exceptions \cite{hahn2022crosslinguistic}. 
This study is 
the first to introduce an explicit model designed to disentangle orthogonal mechanisms that work to make linguistic systems communicatively optimal. 
These include forces that govern births and losses of individual word forms, as well as pressures that introduce forms into and remove them from salient meaning roles; 
additionally, 
the model employed sheds light on 
processes that mutate word forms over the course of their lifetimes as well as when they occupy 
salient meaning roles. 
Results indicate that different evolutionary mechanisms impact on the introduction and maintenance of efficient patterns 
to different degrees. 

Word forms containing sequences of identical consonants, a characteristic shown to pose problems for word production and comprehension, arise far less frequently than those without, and far less frequently than would be expected under a neutral process of word form generation. 
Forms with identical consonants enter languages' basic vocabularies less frequently than those without, though not at a rate greater than would be expected from a random sample from a language's vocabulary for all but one family analyzed. 
Over the course of word forms' character histories, processes of word form mutation --- a composite of regular sound changes, analogical changes, and in some cases changes to a word family that alter transparent morphological relationships --- are more likely to remove sequences of identical consonants than they are to introduce them; however, this effect is not consistently greater than what would be expected under a neutral process of sound change (as operationalized here), and this effect is not detectable within the durations of time when forms serve as basic vocabulary items for most language families under study.  
Forms with identical consonants are phased out of languages' basic vocabularies more frequently than those without; however, individual word forms with this pattern (serving in any meaning function) 
do not die out more frequently than forms without sequences of identical consonants. 
Word forms containing identical consonants can persist for a long time, likely in less salient meaning functions, 
despite their communicatively suboptimal characteristics. 
Importantly, while competition between forms to fill various niches in the basic vocabulary favors forms without identical consonants \cite{martin2007evolving}, there is no evidence that word forms with this pattern are more likely to die out together, as claimed by some \cite{Frischetal2004}. 
This dovetails with theories in which language change is not driven solely by cognitive considerations \cite{berg1998linguistic,bickel2015neurophysiology}; there may be social pressures 
that favor the retention of lexical items, regardless of the sound patterns they contain. 

This finding 
is all the more intriguing
when 
viewed through the 
lens of evolutionary and developmental perspectives on phenotypic evolution, which tease apart developmental constraints and sorting-related processes in evolutionary trajectories \cite{amundson1994two,fusco2001many}. 
Developmental constraints involve limitations on the space of variants that can be produced, while variants are propagated through sorting-related processes. 
It is not clear whether the object of investigation of this study, lexical items, is directly analogous to biological notions of a phenotype, as human language itself is argued to be a phenotype \cite{reyes2016tracking,Bickeletal2023LanguageFollowsUnique}. 
Regardless, an important insight is that the evolutionary trajectories of lexical items can be shaped by production biases (i.e., constraints on the creation of certain types) as well sorting-related pressures such as selection or drift (here, mutational processes changing the sound patterns found in word forms as well as extinction processes affecting word forms with different sound patterns). 
The cross-linguistic under-representation of word forms containing identical consonants is overwhelmingly due to a bottleneck in production. 
To a less pronounced extent not found across all families studied, mutational processes favor the removal of sequences of identical consonants within word forms over their lifetimes. 
Given the ambiguous nature of these results, we are not in a position to characterize this mutational asymmetry as a selectional process driven by sound changes that explicitly target sequences of identical consonants rather than an epiphenomenon of other change types that happened to remove such sequences in the course of affecting a particular sound in a wider range of contexts (as in the case of Latin {\it bibere} $>$ Old French {\it beivre}). 
Furthermore, the method used here detects asymmetric pressures in evolution but may not reliably distinguish between signatures of drift and selection \cite{reali2010words,newberry2017detecting}. 
Crucially, after surviving the production bottleneck and being subjected to mutational processes, forms with identical consonants have as much longevity as those without, even though they appear to be selected against in the basic vocabulary. 

Despite the general flexibility human languages have in assigning forms to meaning functions \cite{watson2022optionality}, the distribution of sound patterns in languages' lexicons are still shaped by a drive towards communicative efficiency. However, there are multiple components to this general pressure with roles that remain poorly understood. Phylogenetic approaches like the one employed here have the potential to shed light on the origins and maintenance of other plausibly efficiency-driven phenomena \cite{zipf1949human,dautriche2017words}. 










%% file: methods.tex
\subsection{Cognate class traits}
The evolution of cognate classes was analyzed in the Austronesian, Semitic, and Uralic families using data from digitized etymological dictionaries \cite{blust2013austronesian,robert_blust_2023_7741197,list2022lexibank,SEDonline,uralothek,uralonet}. 
These resources organize etymologically related forms in contemporary languages according to cognate classes and provide a reconstructed etymon (i.e., ancestral form) for each cognate class. 
For each language in the three data sets, cognate classes were coded according to whether or not they were absent or present (e.g., Latin {\it manducare} `chew' survives into French as {\it manger} `eat' but has been lost in Spanish), and if present, whether it contained two adjacent (i.e., separated by a vowel) identical consonants or not. 
This yields three states that a language can express for a given cognate class: {\sc absent}, $+$IC, and $-$IC.

The search for identical consonants was restricted to sequences which co-occurred within and not across active morpheme boundaries (e.g., boundaries between members of complex words such as compounds), since a number of key generalizations regarding identical consonant avoidance make reference to tautomorphemic violations of this constraint \cite{greenberg1950patterning,herdan1962patterning,frajzyngier1979notes,vernet2011semitic}. 
In Austronesian languages in particular, co-occurrence rates of consonants with identical place of articulation differ across tautomorphemic and heteromorphemic contexts, given the frequent occurrence of reduplication and infixation processes that create identical adjacent consonants in derived forms \cite{racz2016gradient,zuraw2009diverse}. 
Accordingly, models may infer different degrees of diachronic tolerance for identical consonants, depending on whether only tautomorphemic sequences are taken into consideration. 

In the Semitic and Uralic data sets, hyphens were taken to mark active morpheme boundaries in words where they were present. Detecting synchronically active morpheme boundaries was a greater challenge for the Austronesian data, as the Austronesian Comparative Dictionary (ACD) marks affix and infix boundaries that were active in ancestral forms but not necessarily active in the reflexes where they are marked. 
As an example, the ACD gives the Aklanon word for `woman' as {\it ba-b\'{a}yi} on the basis of reduplicated Proto-Austronesian {*ba-bahi}, even though a morpheme boundary is not marked in the source from which the word is taken \cite{zorc1969} and the form is presumably synchronically tautomorphemic, as there are no other related forms that would facilitate the abstraction of a base {\it b\'{a}yi}. 
Coding only the presence of identical consonants within hyphen-delimited forms after stripping out infixes runs the risk of severely under-counting tautomorphemic violations of IC avoidance. 
In order to address this issue, for a group of etymologically related forms in a given language sharing a transparent semantic relationship and a clear derivational relationship (e.g., Javanese {\it ni\d{t}ik} `to strike a light using flint and steel' and {\it \d{t}i\d{t}ik} `flint and steel for starting fires' $<$ PAN {*tiktik}), the longest common subsequence was extracted (here {\it i\d{t}ik}) and treated as the basic reflex of the etymon in question. 

Each reconstructed etymon in each dataset was aligned with the portion (in the case of Semitic and Uralic, hyphen-delimited morphemes, and in the case of Austronesian, the longest common subsequence found across reflexes, see above) of each corresponding entry most likely to descend from it using an iterative version of the Needleman-Wunsch algorithm \cite{NeedlemanWunsch1970,Jaeger2013}. 
The purpose of this was to minimize the risk of extracting the presence of identical consonants in an element not homologous with the etymon whose evolution is being tracked.

Aligned forms were orthographically normalized in order to facilitate the extraction of identical consonants separated by a single vowel. 
Digraphs corresponding to single sound segments were identified in a variety of ways. 
Some languages are presented in their standard orthography, making it straightforward to identify and modify digraphs corresponding to single segments. 
Strings corresponding to aligned forms were split and potential ligatures combining with other strings were identified from the resulting characters. For each aligned form, clusters of identical segments and ligatures were merged into space-delimited sequences of characters representing a single segment. 
This had the effect of ensuring that digraphs were not treated as sequences of different segments, and also that geminate consonants were treated as a single consonantal unit. 
Sequences of geminate (i.e., doubled) consonants were simplified 
so that they could be identified with their singleton counterparts. Following these processing steps, it was straightforward to automatically extract whether identical consonants separated by a single vowel were present in a form via a script. 
This made it possible to code each cognate class according to the states \{{\sc absent}, {\sc $-$IC}, {\sc $+$IC}\} in each language. 
In some cases, a cognate class attests both of the states {\sc $-$IC} and {\sc $+$IC} in a given language, if the language attests synchronically unrelated reflexes of an etymon with and without identical consonants. 

A final processing step for phylogenetic analysis was to convert the data sets into likelihood matrices, setting state values for a given etymon in a given language to 1 and all unattested values to 0. 
Since languages often attest more than one value for a given etymon, some languages had multiple likelihoods set to one for different etyma. 
It is worth highlighting that this is a method for dealing with data ambiguity in cladistics rather than actual polymorphism \cite{Felsenstein2004}. 
For phylogenetic comparative analyses conducted on these data sets, published tree samples of the Austronesian, Semitic and Uralic families were used \cite{gray2009language,kitchen2009bayesian,honkola2013cultural}. 

To ensure that well-etymologized languages and secure cognate classes were used for analyses, datasets contain only languages with more than 250 reflexes in the etymological database in which they are found and cognate classes found in more than 10\% of languages in a given family. 
The Austronesian data set consisted of 1693 cognate sets from 54 languages. 
The Semitic data set consisted of 1378 cognate sets in 23 languages. 
The Uralic data set consisted of 1872 cognate sets in 15 languages. 

\subsubsection{Phylogenetic analysis of cognate class traits}

Cognate class traits were assumed to evolve over phylogenies according to a continuous-time Markov (CTM) chain, a stochastic process where between-state transitions take place according to transition rates. 
A number of biological studies have used CTM models to analyze the evolution of morphologically dependent traits, such as tail color, which is only relevant if a tail is present in a species \cite{Maddison1993,tarasov2019integration}. 
A crucial difference between biological phenomena of this sort and cognate class traits is that cognate classes are non-homoplastic; they are generally born once on a phylogeny (except in the case of extensive borrowing or parallel derivational processes), and cannot be revived once they die out, in the absence of a strong philological tradition similar to that of contemporary times.

In order to ensure that the evolutionary model used has the single-birth behavior described above, I use a modified version of the Stochastic Dollo model of character evolution \cite{nicholls2006quantifying,alekseyenkoetal2009} that does not suffer from well-known problems of this method, 
in that the initial character state is independent of the character's long-term behavior, and the likelihood of $D$ attested cognate classes under a phylogeny $\Psi$ and evolutionary rate parameters $\boldsymbol{Q}$, $\prod_{d=1}^D P(\boldsymbol{x}_d|\Psi,\boldsymbol{Q})$ can be efficiently computed using the standard pruning algorithm \cite{Felsenstein1981}. 
The model used in this paper satisfies the single-birth criterion 
by allowing transitions from the state {\sc absent} to the states $\pm$IC but not from the states $\pm$IC to the state {\sc absent} on potential birth loci, i.e., branches ancestral to the most recent common ancestor (MRCA) of all languages where the cognate class is present, and from $\pm$IC to {\sc absent} but not {\sc absent} to $\pm$IC on all other branches. This ensures that a cognate class will be born once on a phylogeny, and not be revived once it dies out. 

Since the reconstructions found in the etymological resources used were arrived at by experts via careful application of the comparative method of historical linguistics, care was taken to ensure that the initial state ($\pm$IC) of each cognate class character matched the presence or absence of identical consonants in the reconstructed form. 
This involved grafting a branch of infinitesimal length to the MRCA of all languages where the cognate class is present leading to a node containing the state found in the expert reconstruction. 
Additionally, transitions between the states $\pm$IC were not allowed on birth loci, ensuring that the birth state of each cognate class matched the state found at the tip of the grafted branch. 

For an individual cognate class with index $d \in \{1,...,D\}$, transitions between the states \{{\sc absent}, {\sc $-$IC}, {\sc $+$IC}\} take place according to the following rate matrix on birth loci
(diagonal cells are equal to the negative sum of off-diagonal cells in the same row):
$$
Q^{b}_d = \left ( \begin{matrix}
  \text{---} & \lambda^{-}_d & \lambda^{+}_d \\
   0 & \text{---} & 0 \\
   0 & 0 & \text{---} \\
\end{matrix}
\right )
$$
On non-birth loci, the rate matrix takes the following form:
$$
Q^{\neg b}_d = \left ( \begin{matrix}
  \text{---} & 0 & 0 \\
   \mu^{-}_d & \text{---} & \rho^{-+}_d \\
   \mu^{+}_d & \rho^{+-}_d & \text{---} \\
\end{matrix}
\right )
$$
The birth rate parameters $\lambda^{-}_d$ and $\lambda^{+}_d$ represent 
transitions from the state {\sc absent} to the states {\sc $-$IC} and {\sc $+$IC}, respectively; $\rho^{-+}_d$ and $\rho^{+-}_d$ represent transitions between the states {\sc $-$IC} and {\sc $+$IC}; and $\mu^{-}_d$ and $\mu^{+}_d$ represent transitions from the states {\sc $-$IC} and {\sc $+$IC}, respectively, to the state {\sc absent}. 
As in other modifications to the Stochastic Dollo model \cite{bouckaert2017pseudo}, cognate traits cannot be born again after they have been active and lost.

Since cognate classes are born only once, the birth rates $\lambda^{-}$ and $\lambda^{+}$ are kept invariant across cognate classes. 
The remaining evolutionary parameters, which pertain to the evolution of cognate classes once they are born, are allowed to vary according to a hierarchical model for each cognate class $d \in \{1,...,D\}$, since individual cognate classes may have different evolutionary trajectories. 
According to this model, cognate class-specific transition rates are composed of a global rate and a local cognate class-specific multiplier that allows rates to vary across classes as needed. 
Rates are distributed as described below.

Priors over 
the parameters 
$\lambda^{-}_0$, 
$\lambda^{+}_0$, 
$\rho^{-+}_0$, 
$\rho^{+-}_0$, 
$\mu^{-}_0$, 
$\mu^{+}_0$, 
which represent log mean rates around which cognate class-level rates are distributed, 
follow the standard normal distribution. 
For a given cognate class with index $d \in \{1,...,D\}$, evolutionary rates have the following form. 
The global birth rates are transformed via an exponential link function:
\begin{description}
    \item $\lambda^{-}_d = \exp{(\lambda^{-}_0)}$
    \item $\lambda^{+}_d = \exp{(\lambda^{+}_0)}$
\end{description}
The remaining transition rates are log-normally distributed:
\begin{description}
    \item $\rho^{-+}_d \begin{cases}
        \sim \text{LogNormal}(\rho^{-+}_0,\sigma^{\rho^{-+}}) & \text{if } \boldsymbol{x_d} \in \{\text{\sc absent},\text{$-$IC},\text{$+$IC}\} \\
        = 0 & \text{otherwise}
    \end{cases}$
    \item $\rho^{+-}_d \begin{cases}
        \sim \text{LogNormal}(\rho^{+-}_0,\sigma^{\rho^{+-}}) & \text{if } \boldsymbol{x_d} \in \{\text{\sc absent},\text{$-$IC},\text{$+$IC}\} \\
        = 0 & \text{otherwise}
    \end{cases}$
    \item $\mu^{-}_d \sim \text{LogNormal}(\mu^{-}_0,\sigma^{\mu^{-}})$
    \item $\mu^{+}_d \sim \text{LogNormal}(\mu^{+}_0,\sigma^{\mu^{+}})$
\end{description}
$\text{HalfNormal}(0,1)$ priors are placed over standard deviation parameters $\boldsymbol{\sigma}$. 
Not all cognate classes attest all three states; some only express the pairs of states ({\sc abs},$-$IC) and ({\sc abs},$+$IC). 
These characters do not provide information that bears on transitions between the states $-$IC and $+$IC, but provide information regarding the birth rates and death rates of cognate classes displaying these patterns. 
For characters of this sort, transitions to and from the unattested state are set to zero, as shown above. 

The likelihood of each trait $P(\boldsymbol{x}_d|\Psi,\boldsymbol{Q_d})$ was corrected for ascertainment bias. 
This correction is intended to account for the fact that the observed cognate classes represent only a fraction of the cognate classes that have existed during the course of each family's history, as many will have died out before being recorded \cite{felsenstein1992phylogenies,Felsenstein1981,Bouckaertetal2012,Changetal2015}. 
This amounts to conditioning the trait likelihood on the probability that the trait would be observed in the first place under the CTM process that governs its evolution. 
The corrected likelihood is equal to the following:
$$\frac{{P(\boldsymbol{x}_d|\Psi,\boldsymbol{Q_d})}}{{1-P(\boldsymbol{x}_{\text{\sc abs}}|\Psi,\boldsymbol{Q_d})}}$$
Above, $\boldsymbol{x}_{\text{\sc abs}}$ represents a trait likelihood matrix 
with the value {\sc absent} 
for all tips in the phylogeny. 
For comparability between $P(\boldsymbol{x}_d|\Psi,\boldsymbol{Q_d})$ and $P(\boldsymbol{x}_{\text{\sc abs}}|\Psi,\boldsymbol{Q_d})$, $\boldsymbol{x}_{\text{\sc abs}}$ 
is augmented to contain a tip descending from a branch of infinitesimal length grafted to the MRCA of all languages where the cognate class is present, the value of which is equal to the reconstructed value.

\subsubsection{Baselines for cognate class traits}

\paragraph{Baseline birth rates of cognate class traits}

Under a process where sequences are generated by randomly sampling consonants from the uniform distribution, the probability of generating a sequence $w$ containing at least two adjacent identical consonants is equal to the following, where $|w|$ denotes sequence length and $+\text{IC} \in w$ indicates the presence of adjacent identical consonants within a sequence:
$$
P(+\text{IC} \in w) = \sum_{i=1}^{N} P(|w|=i) P(+\text{IC} \in w; |w|=i) 
$$
In a language with $S$ segments, $P(+\text{IC} \in w; |w|=2) = \frac{1}{S}$. 
Since the probability of generating a sequence containing at least two adjacent identical consonants is higher for longer sequences, $P(+\text{IC} \in w)$ will be higher when $P(|w|=i)=\frac{1}{N}$ for all $i \in \{1,...,N\}$. 
Assuming that shorter sequences are more frequently generated than longer ones, we expect this quantity to approach $\frac{1}{S}$ as $P(|w|=2)$ approaches $1$, 
allowing us to derive a lower bound $P(+\text{IC} \in w) \geq \frac{1}{S}$. The expected ratio between words without and words with identical consonants will then be less than or equal to $S-1$. 
In the case of a theoretical language requiring that a minimal word consist of more than two consonants, this ratio will be even smaller. Numbers of consonants for languages in each family (Afro-Asiatic was taken as a proxy for Semitic) were taken from the PHOIBLE database \cite{phoible}. 

\paragraph{Baseline $+$IC $\rightarrow$ $-$IC vs.\ $-$IC $\rightarrow$ $+$IC mutation rates}

A simulation procedure was used to estimate the frequencies at which 
neutral models of sound change are expected to introduce sequences of identical consonants into lexical items versus remove them. 
Frequencies of such changes depend on existing frequencies of sound patterns found across the lexicon. To ensure that frequencies of word lists to which simulated sound changes were realistic, word lists from languages in each data set were used (simulations were applied to languages with 500 or more entries). 
For each language, an input segment type was chosen at random from the language's inventory. The type of change --- (1) whether it was unconditioned, i.e., affecting all segments of a particular type, or affected only (2) word-initial or (3) word-medial segments --- was chosen at random. Finally, an output segment was chosen from the language's inventory at random; for changes affecting word-initial or word-medial segments, deletions were also allowed. 
After converting the input segment to the output segment in the relevant environment (depending on the change type) across the lexicon, the number of $+$IC $\rightarrow$ $-$IC vs.\ $-$IC $\rightarrow$ $+$IC changes were tabulated, and a ratio computed by dividing the former number by the latter number (with a small constant added to each number in the case of zero division). This procedure was carried out 5000 times per language, with ratios averaged at the language level.

\subsection{Cognate-concept traits}

The evolution of cognate-concept (alternatively root-meaning) traits \cite{Ringeetal2002,Nakhlehetal2005,Changetal2015} was analyzed using data from a subset of the Lexibank repository \cite{johann_mattis_list_2021_5227818} that has been further processed to normalize orthographic forms as well as link forms in different languages to the Concepticon semantic taxonomy \cite{concepticon}. 
I used data sets for which cognacy was coded and for which reliable phylogenetic tree samples have been published. 
Data from five families were analyzed. 
These were 
Dravidian \cite{kolipakam2018dravlex,kolipakam_vishnupriya_2021_5121580,kolipakam2018bayesian}, 
Indo-European \cite{dunn2012indo,dunn_michael_2021_5121651,Changetal2015}, 
Sino-Tibetan \cite{Sagartetal_database_2019,sagart2019dated}, 
Turkic \cite{savelyev2020bayesian,alexander_savelyev_2021_5137274}, 
and 
Uto-Aztecan \cite{greenhill2023recent}. 

Forms in different languages were automatically coded according to whether or not they contained a sequence of identical consonants separated by a single vowel within morpheme boundaries (demarcated by the symbol $+$). 
This was relatively straightforward thanks to the space-delimited orthographic normalization of forms. 
The Cross-Linguistic Transcription Systems (CLTS) database \cite{anderson2018cross} was used to determine which segments in each string were consonants. The geminate marker {\IPA :} was stripped from geminate segments and sequences of identical segments were simplified to one segment before a script was used to  detect the presence of adjacent identical consonants within morphological boundaries.


A language expresses a given semantic concept using formal material corresponding to one or more cognate classes, in which sequences of identical consonants can be present or absent. 
For instance, Portuguese expresses the concept DRINK with the form {\IPA /b\textbari beR/}, which 
contains identical consonants and is a reflex of the Proto-Indo-European etymon {*peh$_3$-}. 
Thus, for each language in a family, cognate-concept traits are coded according to the states \{{\sc absent},$-$IC,$+$IC\}. 

Cognate-concept characters for different families were transformed into binarized likelihood matrices. In the case of lexical polymorphism (i.e., in which a language attests multiple forms for a meaning), multiple likelihoods were set to one. 
Analyses were restricted to data corresponding to 100 basic concepts \cite{Swadesh1955} available through Concepticon \cite{concepticon}. 
Concept rankings were taken from NorthEuraLex \cite{dellert2020northeuralex}. 
The Dravidian data set consisted of 
709 concept-cognate traits corresponding to 93 concepts from 20 languages. 
The Indo-European data set consisted of 
686 concept-cognate traits corresponding to 96 concepts from 19 languages. 
The Sino-Tibetan data set consisted of 
1517 concept-cognate traits corresponding to 83 concepts from 44 languages. 
The Turkic data set consisted of 
225 concept-cognate traits corresponding to 90 concepts from 31 languages. 
The Uto-Aztecan data set consisted of 
1087 concept-cognate traits corresponding to 92 concepts from 33 languages.


\subsubsection{Phylogenetic analysis of cognate-concept traits}
Cognate-concept traits were modeled as evolving according to a CTM process. 
Since they are homoplastic (i.e., a cognate class can come to express the same meaning independently on two different lineages), standard models used to analyze morphologically dependent traits are applicable without the need to account for the single-birth criterion. 

As above, hierarchical models were used to jointly analyze the evolution of cognate-concept traits jointly within separate families. 
Transition rates were assumed to vary at the concept level; the likelihood for a given cognate-concept trait with index $d \in \{1,...,D\}$ under a phylogeny $\Psi$, $P(\boldsymbol{x}_d|\Psi,Q_{\text{concept}[d]})$ depends on the transition rates for the concept which the trait expresses 
and can be computed using the pruning algorithm. 

For each concept $c \in \{1,...,C\}$, 
transitions between the states \{{\sc absent}, {\sc $-$IC}, {\sc $+$IC}\} take place according to the following rate matrix:
$$
Q_c = \left( \begin{matrix}
  \text{---} & \lambda_c^{-} & \lambda_c^{+} \\
  \mu_c^{-} & \text{---} & \rho_c^{-+} \\
  \mu_c^{+} & \rho_c^{+-} & \text{---} \\ 
\end{matrix} \right)
$$
Here, all rates (including the birth rates $\lambda^{-}$ and $\lambda^{+}$) vary across concepts, since cognate-concept traits are homoplastic, and concept-cognate traits for certain concepts may arise more frequently than for others. 

Priors over 
the parameters 
$\lambda^{-}_0$, $\lambda^{+}_0$, $\rho^{-+}_0$, $\rho^{+-}_0$, $\mu^{-}_0$, $\mu^{+}_0$, 
which represent log baseline rates, 
follow the standard normal distribution. 
For a given concept with index $c \in \{1,...,C\}$, evolutionary rates are distributed as follows:
\begin{description}
    \item $\lambda^{-}_c \sim \text{LogNormal}(\lambda^{-}_0,\sigma^{\lambda^{-}})$
    \item $\lambda^{+}_c \sim \text{LogNormal}(\lambda^{+}_0,\sigma^{\lambda^{+}})$
    \item $\rho^{-+}_c \sim \text{LogNormal}(\rho^{-+}_0,\sigma^{\rho^{-+}})$
    \item $\rho^{+-}_c \sim \text{LogNormal}(\rho^{+-}_0,\sigma^{\rho^{+-}})$
    \item $\mu^{-}_c \sim \text{LogNormal}(\mu^{-}_0,\sigma^{\mu^{-}})$
    \item $\mu^{+}_c \sim \text{LogNormal}(\mu^{+}_0,\sigma^{\mu^{+}})$
\end{description}
$\text{HalfNormal}(0,1)$ priors are placed over standard deviation parameters $\boldsymbol{\sigma}$. 
The rate parameters for concept-cognate trait $d \in \{1,...,D\}$ are equal to the rate parameters for $\text{concept}[d]$, if $\boldsymbol{x}_d$ attests all three states \{{\sc absent},$-$IC,$+$IC\}; otherwise, $\rho^{-+}_d$ and $\rho^{+-}_d$ are set to zero, as in the previous study.

Trait likelihoods were corrected for ascertainment bias in the manner described above. Here, $\boldsymbol{x}_{\text{\sc abs}}$ represents a trait likelihood matrix 
with the value {\sc absent} 
for all tips in the phylogeny. 

\subsubsection{Baselines for cognate-concept traits}

\paragraph{Baseline birth rates of cognate class traits}

Under a null model in which basic vocabulary items are sampled from the general (i.e., basic and nonbasic) vocabulary at random, with no sensitivity to the sound patterns displayed by individual forms, the ratio of birth rates of cognate-concept traits without versus with sequences of identical consonants should be comparable to the ratio between forms without and with identical consonants in the lexicon from which basic vocabulary items are sampled. 

These ratios are estimated for languages in each family under study on the basis of large word lists comprising basic as well as non-basic items. Dravidian, Indo-European and Turkic ratios were estimated from NorthEuraLex \cite{dellert2020northeuralex}. Sino-Tibetan ratios were estimated from the Sino-Tibetan Etymological Dictionary and Thesaurus \cite{matisoff2015sino}. 
Uto-Aztecan ratios were estimated from available digitized resources for Nahuatl \cite{ids-222}, Yaqui \cite{wold-32} and the Bridgeport dialect of Northern Paiute \cite{northern-paiute}. 
For each language, the number of forms lacking sequences of identical consonants was divided by the number of forms containing sequences of identical consonants.

\paragraph{Baseline $+$IC $\rightarrow$ $-$IC vs.\ $-$IC $\rightarrow$ $+$IC mutation rates}

This simulation procedure was carried out as described for cognate class traits, with the difference that sound changes were applied only to the 100 basic vocabulary items under analysis rather than larger word lists.

\subsection{Inference}

Data were processed using Python 3 as well as version 0.6-99 of the R package {phytools} \cite{phytools}. 
Models were fitted using RStan version 2.26.13 \cite{Carpenteretal2017}, running the No U-Turn Sampler (NUTS) over 4 chains for 2000 iterations, with the first half discarded as burn-in. 
Model convergence was assessed via the potential scale reduction factor \cite{GelmanRubin1992}, with values under $1.1$ taken to indicate convergence. 
To incorporate phylogenetic uncertainty, the model was run on 25 trees from each tree sample and the resulting posterior samples for runs that reached convergence were concatenated together, yielding 100000 samples per model. 
95\% HDIs were computed using the R package {HDInterval} \cite{HDI}.
Data and code used can be found at \url{https://github.com/chundrac/idcc}. 

%% file: acknowledgements.tex
This work was supported by the DFG Center for Advanced Studies -- Words, Bones, Genes, Tools, University of T\"ubingen and the NCCR Evolving Language (Swiss National Science Foundation Agreement No. 51NF40\underline{\phantom{X}}180888). 
SNSF grant No.\ 207573 is gratefully acknowledged as well. 
I thank audiences in Potsdam, T\"ubingen, and Copenhagen for thoughtful feedback. This work 
was greatly improved due to 
conversations with Gerhard J\"ager, Adamantios Gafos, and Marina Laganaro. 
Johannes Dellert drew my attention to the need for more principled baselines against which to evaluate hypotheses, as did Erich Round, who provided extensive guidance regarding how to derive such baselines along with other valuable feedback. 
I am grateful to Will Bausman and Marcel Weber for highlighting points of contact with the Evo-Devo framework. 
All errors and infelicities are my own.